\PassOptionsToPackage{dvipsnames}{xcolor}
\documentclass[11pt,letterpaper]{style}

\usepackage[numbers]{natbib}
\usepackage{graphicx}
\usepackage{booktabs}
\usepackage{amsmath,amsfonts,amssymb}
\usepackage{cleveref}
\usepackage{subcaption}
\usepackage{wrapfig}
\usepackage{multirow}
\usepackage{colortbl}
\usepackage{listings}
\usepackage{xparse}
\usepackage{fontawesome5}
\usepackage{bxcoloremoji}
\usepackage{float}
\usepackage{placeins}
\usepackage{threeparttable}

\graphicspath{{./}{fig/}{figures/}{plot/}{pdf/}{table/}}
\usepackage{amsthm}
\usepackage{tcolorbox}
\usepackage{svg}
\tcbuselibrary{skins,breakable}
\tcbuselibrary{listingsutf8}
\usepackage{titletoc}

\usepackage{setspace}
\usepackage{pifont}
\usepackage{mathtools}
\usepackage{enumitem}
\usepackage{arydshln}
\usepackage{bbm}
\usepackage{lineno}
\usepackage{makecell}
\usepackage{adjustbox}
\usepackage{algorithm}
\usepackage{algorithmic}
\usepackage{caption}
\usepackage{hyperref}
\usepackage{bbding}
\usepackage{xspace}

\usepackage[utf8]{inputenc} 
\usepackage[T1]{fontenc}    
\usepackage{hyperref}       
\usepackage{url}            
\usepackage{booktabs}       
\usepackage{amsfonts}       
\usepackage{nicefrac}       
\usepackage{microtype}      
\usepackage{xcolor}         
\usepackage{graphicx}
\usepackage{amsmath}

\usepackage{xcolor}
\usepackage{tabularray}
\usepackage{wrapfig}
\usepackage[dvipsnames]{xcolor}
\newcommand{\inc}[1]{\textcolor{BrickRed}{#1}}
\newcommand{\dec}[1]{\textcolor{ForestGreen}{#1}}

\definecolor{stateColor}{HTML}{88A7D6}
\definecolor{actionColor}{HTML}{A83A24}
\definecolor{transitionColor}{HTML}{9C75AE}
\usepackage[table]{xcolor}
\usepackage{tabularx}
\usepackage{multirow}
\usepackage{booktabs} 
\usepackage{xcolor}   
\usepackage{makecell} 
\usepackage{pifont}
\usepackage{adjustbox}

\newcommand{\Statecolor}[1]{\textcolor{stateColor}{#1}}
\newcommand{\Actioncolor}[1]{\textcolor{actionColor}{#1}}
\newcommand{\Transitioncolor}[1]{\textcolor{transitionColor}{#1}}
\newcommand{\ours}[1]{\texttt{AndroidReality}}

\usepackage[table]{xcolor}
\usepackage{tabularray}
\UseTblrLibrary{booktabs}
\usepackage{makecell}
\usepackage{graphicx}

\definecolor{backheader}{HTML}{D9EAF7}
\definecolor{backblue}{HTML}{EEF6FF}
\definecolor{backred}{HTML}{FFF0F0}
\definecolor{backpurple}{HTML}{F5F0FF}

\definecolor{statecolor}{HTML}{1F77B4}
\definecolor{actioncolor}{HTML}{D62728}
\definecolor{transitioncolor}{HTML}{6F42C1}

\definecolor{backblue}{HTML}{E5EBF6} 
\definecolor{backred}{HTML}{FAF4F2} 
\definecolor{backpurple}{HTML}{F7F3F8} 
\definecolor{backheader}{HTML}{ECECEC}
\newcommand{\pert}[1]{\begingroup\Urlmuskip=0mu plus 1mu\path{#1}\endgroup}

\definecolor{mygray}{gray}{0.9}
\definecolor{syncol}{RGB}{243,246,249}
\definecolor{wildcol}{RGB}{215,240,235}
\definecolor{drop1}{RGB}{180,225,220}
\definecolor{drop2}{RGB}{150,210,200}
\definecolor{drop3}{RGB}{120,195,185}
\definecolor{drop4}{RGB}{95,180,170}
\definecolor{drop5}{RGB}{65,160,150}
\definecolor{lightblue}{RGB}{26,82,249}

\definecolor{myblue1}{HTML}{0171DC}
\definecolor{myblue2}{HTML}{013978}

\NewDocumentEnvironment{minted}{O{} m +b}{%
}{}

\newcommand{\equalmark}{\textsuperscript{*}}
\newcommand{\corrmark}{\textsuperscript{\dag}}

\title{\ours{}: How Far Are Mobile Agents from the Real World?}
\runningtitle{\ours{}: How Far Are Mobile Agents from the Real World?}

\author{%
    {\Authfont
    Xiaoou Liu\textsuperscript{1}\equalmark \quad
    Longchao Da\textsuperscript{1}\equalmark \quad
    Hanyang Chen\textsuperscript{1} \quad
    Yuan Ling\textsuperscript{2} \quad
    Hua Wei\textsuperscript{1}\corrmark
    }\\
    {\Affilfont
    \textsuperscript{1} Arizona State University \quad
    \textsuperscript{2} Independent Researcher \\
    \texttt{\{xiaoouli, longchao, hchen478, hua.wei\}@asu.edu} \\
    \texttt{ericalingyuan@gmail.com}
    }%
}

\begin{document}


\begin{abstract}
Mobile agents have achieved promising results on clean online benchmarks such as AndroidWorld, yet their performance often degrades sharply in real-world deployment due to environmental variations and imperfect interface conditions. 
In this work, we introduce \textbf{\ours{}}, a perturbation-based framework for evaluating and improving the robustness of mobile agents. 
Through a Markov Decision Process (MDP) perspective, we organize real-world interface variability into a principled taxonomy of perturbations along three axes: state, transition, and action. 
Guided by this taxonomy, we build a perturbed mobile benchmark on top of AndroidWorld with realistic and controllable perturbation injections, enabling systematic robustness evaluation of mobile agents. 
Our evaluation reveals substantial robustness gaps and four recurring error categories, motivating a simple training-free Test-Time Introspective Recovery (TTIR) mechanism that mitigates these failures on both perturbed and clean settings. 
Together, these results position robustness as a missing dimension in mobile agent evaluation and establish benchmark perturbation as an effective tool for both stress testing and surfacing latent weaknesses of mobile agents. Benchmark and implementation are available \href{https://github.com/Xiao0o0o/AndroidReality}{\textcolor{purple}{here}}. 
\end{abstract}

\newcommand{\TitleLinks}{%
\centering
    \vspace{8pt}
}

\maketitle

\begingroup
\renewcommand{\thefootnote}{\fnsymbol{footnote}}
\footnotetext[1]{Equal contribution.}
\footnotetext[2]{Corresponding author.}
\endgroup


\section{Introduction}

\begin{figure}[h!]
    \centering
    \includegraphics[width=\linewidth]{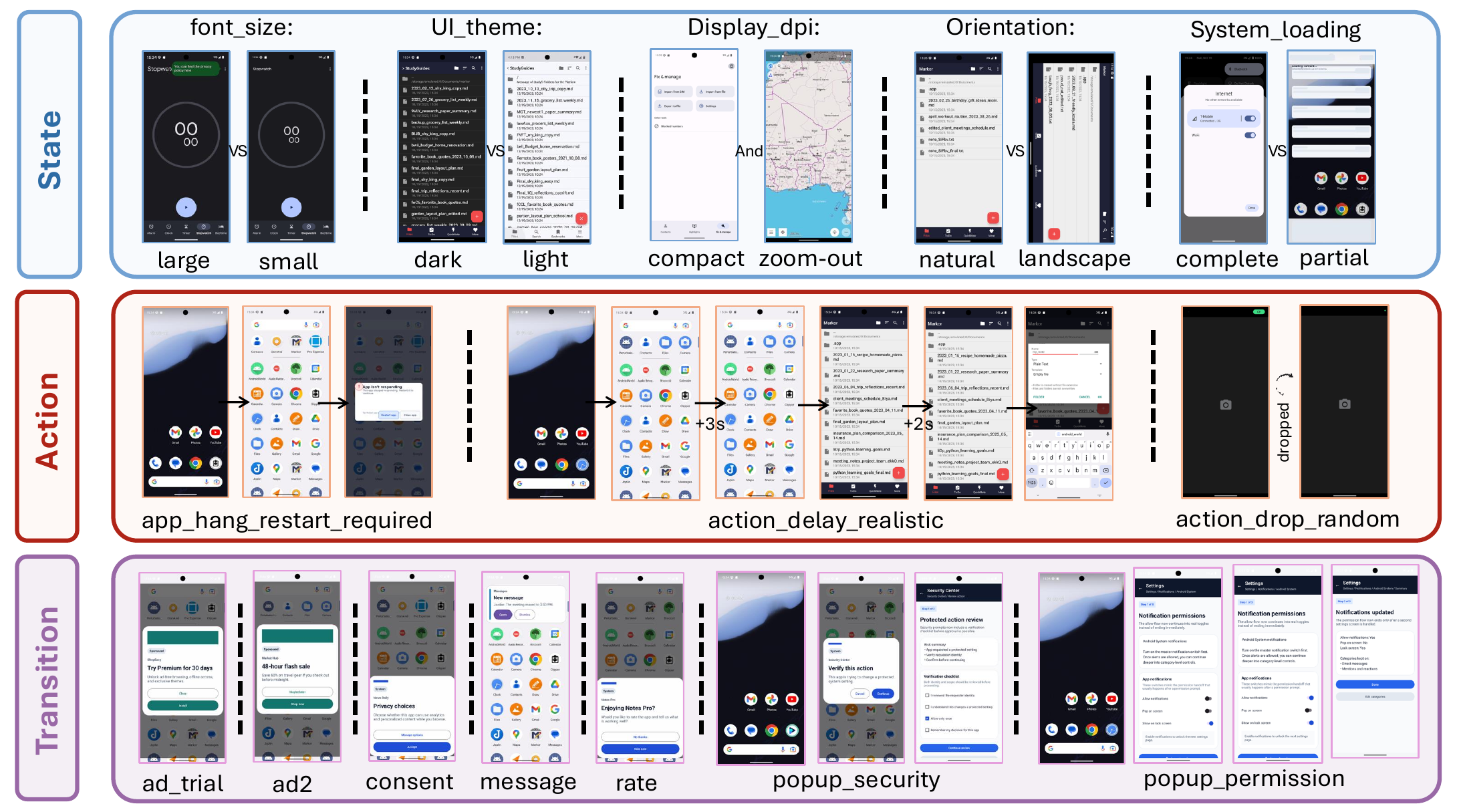}
    \caption{\small Examples of perturbations introduced by \ours{}, designed to emulate real-world deployment conditions that mobile agents face but that existing benchmarks largely overlook. \Statecolor{State-level} perturbations modify the observation surface, including font-size changes, UI theme changes, display-density and screen-size changes, orientation shifts, locale/date-format variations, and partial-loading skeleton screens, etc.  \Actioncolor{Action-level} perturbations alter the execution of the agent's intended actions, including realistic action delays, randomly dropped actions, frozen or loading execution states, app hangs requiring restart, and resets to the home screen.  \Transitioncolor{Transition-level} perturbations insert unexpected intermediate states, including trial offers, interstitial ads, consent sheets, permission dialogs, security pop-ups, rating prompts, notifications, and app/system update prompts.
}
    \label{fig:premain_example}
\end{figure}

Mobile agents~\cite{cheng2025kairos, gu2026generalization, li2025mobileuse, qin2025uitars, ye2025mobileagentv3, xu2026mobileagentv35} have recently achieved strong performance on online evaluation environments~\cite{chen2024spa, kong2025mobileworld, rawlesandroidworld, xu2025androidlab}, demonstrating promising capabilities in multi-step GUI grounding, planning, and action execution. 
These benchmarks played an important role in accelerating progress by providing standardized interfaces and reproducible tasks. 
However, despite encouraging results in such well-defined and controlled settings, the performance of current agents often degrades substantially in real-world deployment~\cite{li2025mobileuse, kong2025mobileworld, wu2024foundations, yang2026gui}. 
As shown in Figure~\ref{fig:premain_example}, practical mobile environments expose agents to a wide range of conditions that are absent from existing benchmarks, including transient pop-ups and notifications, delayed UI responses, varying display configurations, and unreliable action execution, which create a substantial gap between benchmark performance and real-world robustness~\cite{chen2026d, zhang2025hyperclick}.

This gap suggests that current benchmark success may overestimate agent reliability under the variability of real-world deployment. 
Clean benchmarks instantiate a simplified and stationary environment: the interface follows a canonical layout, the environment transitions predictably in response to actions, and action execution is assumed to be reliable. 
Real-world deployment, by contrast, deviates from each of these assumptions along well-defined axes, namely interface presentation, environment dynamics, and action execution~\cite{da2025survey, long2025survey}, and the same task can therefore unfold along many different trajectories. 
As a result, failures in deployment often do not stem from a lack of task-solving ability alone, but from brittleness to environment variations that current agents were never meaningfully exposed to during training or evaluation~\cite{li2025mobileuse, gonzalez2026reliability, luo2026lost}.

In this work, we introduce a perturbation-based framework \ours{} for systematically studying the robustness of mobile agents in online mobile environments. 
Our key idea is to augment a clean benchmark with controlled perturbations that emulate common real-world variations, and to use these perturbations to systematically characterize where and why current mobile agents fail.
We adopt a Markov Decision Process (MDP) view of mobile interaction, where a task is modeled as 
$
\mathcal{M} = (\mathcal{S}, \mathcal{A}, \mathcal{T}, \mathcal{R}, \gamma),
$
with $\mathcal{S}$ the space of UI states, $\mathcal{A}$ the GUI action space, $\mathcal{T}$ the transition dynamics induced by the operating system and target applications, and $\mathcal{R}$ the task completion signal. 
Real deployment can be naturally viewed as a family of perturbed MDPs that share task semantics with the canonical benchmark MDP but vary along specific components of the MDP tuple, and robustness, under this view, is the ability of an agent to maintain task performance across this family.

This formulation suggests a natural taxonomy: perturbations can be organized by which component of the MDP they act upon. 
\Statecolor{State-level} perturbations modify the surface form of state observations without altering the underlying transition structure (e.g., font size changes, orientation shifts, etc.). 
\Transitioncolor{Transition-level} perturbations modify the environment dynamics $\mathcal{T}$ by inserting unexpected intermediate states or redirecting action outcomes (e.g., pop-up interruptions, system notifications, etc.). 
\Actioncolor{Action-level} perturbations modify the mapping between intended and executed actions (e.g., delays, dropped actions, and unintended actions). 
This taxonomy provides a unified foundation for organizing the perturbation space and for attributing robustness failures to specific stages of the agent-env interaction.

Guided by this taxonomy, we construct \ours{}, an online perturbed mobile benchmark that injects realistic and parameterizable variations along all three axes flexibly. 
While prior robustness-oriented benchmarks have focused on individual subsets of these axes, such as naturalistic pop-ups~\cite{yang2026gui, chen2026d} or task-template variation~\cite{gu2026generalization}, \ours{} provides more comprehensive coverage of mobile interface variability through a unified interface. 
Because the perturbations are controlled, they enable fine-grained robustness evaluation beyond aggregate task success rate, allowing us to diagnose which class of perturbation is responsible for the largest share of failures for a given agent. 
Building on this evaluation, we conduct a systematic error analysis that reveals a set of recurring failure patterns across state-of-the-art mobile agents, ranging from distraction by interrupting elements to rigidity in following outdated plans to the inability to recognize and recover from execution failures.

Motivated by these patterns, we introduce a \emph{Test-Time Introspective Recovery} mechanism that monitors the agent's interaction trajectory, detects when it has entered one of the identified failure modes, and triggers a targeted recovery action before continuing with the task. 
The mechanism is training-free and model-agnostic and can be plugged into existing mobile agents without modifying their underlying policy. 
We additionally find that applying this module can recover failure cases in the original clean setting as well, suggesting that the failure modes uncovered through perturbation-based stress testing are not artifacts of perturbation alone but rather reflect broader weaknesses in current mobile agents.
Our contributions can be summarized as follows:

\begin{itemize}
    \item We construct \textbf{\ours{}}, an online mobile benchmark that injects realistic and controllable perturbations into the agent--environment interaction loop, organized by a perturbation taxonomy grounded in the MDP formalism (state, transition, or action). 
    \item We conduct a \textbf{systematic robustness evaluation and error analysis} of SOTA mobile agents on \ours{}, identifying which classes of perturbation drive the largest performance degradation and characterizing four recurring failure patterns behind them.
    \item Guided by this error analysis, we propose a test-time introspective recovery module, a training-free, plug-and-play intervention that can recover some failure cases on both perturbed and clean settings.
\end{itemize}

\section{Related Work}

\subsection{GUI Agent Systems}

Mobile GUI agents have evolved rapidly with the rise of multimodal foundation models~\cite{steiner2024paligemma, xie2024show, bai2025qwen3, team2025kimi, wei2026deepseek}, progressing from rule-based scripts to LLM/VLM-driven autonomous systems~\cite{rawlesandroidworld, wang2024mobile}. Representative agents such as \textit{CogAgent}~\cite{hong2024cogagent} and \textit{AppAgent}~\cite{zhang2024appagent} demonstrate strong planning and grounding capabilities on standard interfaces. However, these agents are primarily developed and evaluated under \emph{clean} interface conditions, where the environment is stationary, and execution is reliable. Recent studies have shown that such agents can be easily distracted by deceptive content~\cite{liu2025hijacking} or fail under under-specified task rules~\cite{chen2026knowu}, raising concerns about their robustness in deployment. \textit{Different from these works}, we do not propose a new agent architecture; instead, we introduce a perturbation-based framework that systematically evaluates existing agents through a principled MDP-grounded taxonomy, attributing failures to specific stages of the `agent and environment' interaction loop.
\vspace{-3mm}
\subsection{Mobile Agent Benchmarks}

Evaluation for mobile agents has shifted from static datasets to dynamic, online environments. Early benchmarks such as \textit{AITW}~\cite{rawles2023aitw} and \textit{AndroidControl}~\cite{li2024effects} relied on offline human demonstrations and could not assess error recovery. \textit{AndroidWorld}~\cite{rawlesandroidworld} established the standard for online evaluation with reproducible state-based verification, while follow-up benchmarks such as \textit{MobileWorld}~\cite{kong2025mobileworld} and \textit{SPA-Bench}~\cite{chen2024spa} extend this paradigm to longer horizons and more complex application workflows.
A more recent line of work begins to examine robustness under interface anomalies. \textit{GUI-Robust}~\cite{yang2026gui} and \textit{D-GARA}~\cite{chen2026d} evaluate agents in the presence of system interruptions, pop-ups, and network failures, while \textit{MobileSafetyBench}~\cite{lee2026mobilesafetybench} focuses on adversarial safety scenarios. \textit{Different from these works}, which treat anomalies as discrete high-level events with task-level success metrics, we ground our perturbations in the MDP formalism and inject controllable variations at the state, transition, and action levels. This allows fine-grained, MDP-stage-level error attribution rather than aggregate success rates, and converts a canonical benchmark into a family of perturbed MDPs for systematic robustness analysis. Table~\ref{tab:benchmark_comparison} summarizes how \ours{} compares to existing benchmarks along these dimensions.

\begin{table*}[t]
\centering
\small
\setlength{\tabcolsep}{3pt}
\resizebox{0.9\textwidth}{!}{%
\begin{tabular}{l c c c c c c c c}
\toprule
\textbf{Benchmark} & \textbf{Plat.} & \textbf{Env.} & \textbf{\#Tasks} & \textbf{Pert. Type} & \textbf{Timing} & \textbf{Attr.} & \textbf{\#P} & \textbf{MDP Stage} \\
\midrule
\multicolumn{9}{l}{\textit{Standard mobile / GUI agent benchmarks}} \\
AndroidWorld~\citep{rawlesandroidworld}    & M   & D & 116  & --     & --      & Coarse & 0  & --  \\
AndroidLab~\citep{xu2025androidlab}            & M   & D & 138  & --     & --      & Coarse & 0  & --  \\
SPA-Bench~\citep{chen2024spa}                  & M   & D & 340  & --     & --      & Coarse & 0  & --  \\
MobileWorld~\citep{kong2025mobileworld}        & M   & D & 201  & --     & --      & Coarse & 0  & --  \\
\midrule
\multicolumn{9}{l}{\textit{Robustness-oriented benchmarks}} \\
GUI-Robust~\citep{yang2026gui}                 & W+D & S & 5318 & Nat     & Pre     & Coarse & 7      & Obs       \\
D-GARA~\citep{chen2026d}                       & M   & D & 152  & Nat     & Run     & Coarse & $\sim$5 & Obs+Trans \\
MobileSafetyBench~\citep{lee2026mobilesafetybench} & M & D & --   & Adv     & Run     & Coarse & --     & Obs       \\
ST-WebAgentBench~\citep{levy2024st}            & W   & D & --   & Adv+Nat & Pre+Run & Coarse & --     & Obs+Act   \\
\midrule
\rowcolor{gray!15}
\textbf{\ours{}} & \textbf{M} & \textbf{D} & \textbf{116} & \textbf{Nat} & \textbf{Pre+Run} & \textbf{Fine (MDP)} & \textbf{24} & \textbf{Obs+Act+Trans} \\
\bottomrule
\end{tabular}%
}
\caption{\small Comparison of GUI/mobile agent benchmarks across robustness evaluation dimensions. 
\textbf{Plat.}: Mobile (M) / Web (W) / Desktop (D). 
\textbf{Env.}: Static (S) / Dynamic online (D). 
\textbf{Pert. Type}: Adversarial (Adv) / Naturalistic (Nat). 
\textbf{Timing}: Pre-task (Pre) / Runtime (Run). 
\textbf{Attr.}: Error attribution granularity. 
\textbf{\#P}: number of distinct perturbation types. 
\textbf{MDP Stage}: Observation (Obs) / Action (Act) / Transition (Trans).
}
\label{tab:benchmark_comparison}
\end{table*}
\section{\ours{} Benchmark}\label{sec:mdp}
\subsection{MDP Perturbation Formulation}

We augment AndroidWorld with a perturbation layer that operates inside the agent--environment interaction loop, converting the original benchmark into a family of perturbed MDPs:
\begin{equation}
    \mathcal{M}_0 
    =
    (\mathcal{S}, \mathcal{A}, \mathcal{T}, \mathcal{R}, \gamma)
    \xrightarrow{\;\;P_{c,\theta}\;\;}
    \widetilde{\mathcal{M}}_{c,\theta}
    =
    (\widetilde{\mathcal{S}}, \widetilde{\mathcal{A}}, \widetilde{\mathcal{T}}, \mathcal{R}, \gamma)
\end{equation}
where $\mathcal{S}$ denotes the UI state space, $\mathcal{A}$ denotes the set of executable mobile actions, $\mathcal{T}$ denotes the environment transition dynamics, $\mathcal{R}$ denotes the task completion reward, and $\gamma$ is the discount factor. 
The perturbation operator $P_{c,\theta}$ wraps the clean interaction loop and transforms the canonical MDP into a perturbed MDP $\widetilde{\mathcal{M}}_{c,\theta}$, where $c \in \{\Statecolor{\textbf{State}}, \Transitioncolor{\textbf{Transition}}, \Actioncolor{\textbf{Action}}\}$ specifies the MDP component being perturbed, and $\theta$ denotes the perturbation type and its configuration. 

In the clean setting, the agent interacts with a single canonical environment: it observes screenshots, UI trees, and available APIs, produces actions such as tapping, scrolling, typing, or selecting, and receives task-completion feedback from the evaluator. 
In contrast, real mobile environments do not correspond to a single fixed MDP. 
A task with the same semantic goal may unfold under different display settings, unexpected intermediate screens, delayed responses, or imperfect action execution. 
All perturbed MDPs preserve the original task semantics and reward definition, but modify the observation surface, transition trajectory, or action execution process. 
This design allows us to stress-test whether an agent has learned robust task-solving behavior or instead relies on brittle assumptions tied to the clean benchmark.

\subsection{Perturbation Injection}
The perturbation wrapper is implemented as an intermediate layer between the Android emulator and the agent. 
The wrapper intercepts the standard interaction loop and selectively applies perturbation operators according to the chosen MDP component $c$ and configuration $\theta$. 
Formally, a clean interaction step follows $s_{t+1} \sim \mathcal{T}(s_t, a_t)$, while the perturbed step under configuration $\theta$ can be written as:
\vspace{-2mm}
\begin{equation}
    s_{t+1} \sim \widetilde{\mathcal{T}}_{\theta}(s_t,\;\phi^{A}_{\theta}(a_t)), 
    \qquad
    \tilde{o}_{t+1} = \phi^{S}_{\theta}(s_{t+1})
\end{equation}
where $\phi^{A}_{\theta}$ perturbs the agent's action before it is executed, $\widetilde{\mathcal{T}}_{\theta}$ replaces the original transition with one that may lead to unexpected intermediate next states, and $\phi^{S}_{\theta}$ perturbs the observation returned to the agent. The agent then conditions its next decision on $\tilde{o}_{t+1}$ instead of $s_{t+1}$. Each operator corresponds to one of the three perturbation classes defined in our taxonomy. We provide a visual illustration of representative perturbations from each class in Figure~\ref{fig:premain_example} and a full summary in Table~\ref{tab:perturbation_taxonomy}.

\textbf{\Statecolor{State-level} perturbations} ($\phi^{S}_{\theta}$) modify the state returned to the agent while preserving the underlying task progress. These perturbations alter the visual or structural form of the observation, such as font size, UI theme, display density, screen orientation, and partially loaded UI content, so that the agent receives perturbed screenshots, UI trees, or API-visible states, while the task goal and success verifier remain unchanged. \textbf{\Transitioncolor{Transition-level} perturbations} ($\widetilde{\mathcal{T}}_{\theta}$) modify the trajectory induced by the environment dynamics. Instead of directly transitioning from the current state to the next targeted state, the environment may pass through an unexpected intermediate state, such as a trial-offer advertisement, a message notification, a consent sheet, a rating prompt, or a security/permission pop-up. 
These perturbations simulate common mobile interruptions that require the agent to recognize off-trajectory states and recover appropriately.
\textbf{\Actioncolor{Action-level} perturbations} ($\phi^{A}_{\theta}$) modify the mapping from the agent's intended action to the action actually realized in the environment. For example, an intended action may be silently dropped, executed only after a noticeable delay, or followed by an app hang requiring a restart. 
These perturbations expose whether the agent can detect failed execution and revise its plan, rather than assuming that every action was successfully executed.
\subsection{Evaluation Protocol}\label{sec:eva}
For each task, we evaluate mobile agents under both the clean AndroidWorld environment and our perturbed \ours{} environments. 
The clean setting measures standard benchmark competence under the canonical MDP $\mathcal{M}_0$, while the perturbed setting measures robustness under the family of perturbed MDPs $\widetilde{\mathcal{M}}_{c,\theta}$. 
Unless otherwise specified, we instantiate one perturbation type at a time so that the performance change can be attributed to a specific perturbation source.

We report the task success rate under the clean setting and under each perturbation category. 
To quantify robustness degradation, we define the category-level performance drop as:
\begin{equation}
    \Delta_c 
    =
    \mathrm{SR}(\mathcal{M}_0)
    -
    \mathrm{SR}(\widetilde{\mathcal{M}}_{c})
\end{equation}
where $\mathrm{SR}(\mathcal{M}_0)$ denotes the success rate in the clean environment and $\mathrm{SR}(\widetilde{\mathcal{M}}_{c})$ denotes the success rate under perturbation category $c$. 
A larger $\Delta_c$ indicates that the agent is more brittle to that class of perturbation. 
When needed, we also compute perturbation-specific degradation $\Delta_{c,\theta}$ to identify which concrete perturbation type causes the largest failure rate.


Beyond binary task success, perturbations may also increase the number of steps an agent needs to complete the task, for example, by requiring the agent to dismiss a pop-up, wait through a loading delay, or recover from a dropped action.  We therefore additionally report the step overhead:
\begin{equation}
    \Delta_{\text{step},c}
    =
    \overline{T}(\widetilde{\mathcal{M}}_{c})
    -
    \overline{T}(\mathcal{M}_0)
\end{equation}
where $\overline{T}(\mathcal{M})$ denotes the average number of steps used by the agent under environment $\mathcal{M}$. 
\section{Experiment}

\subsection{Experiment Setup}
\label{sec:setup}

\noindent \textbf{Benchmark and task environment.}
We evaluate mobile agents on \ours{}, a perturbed benchmark built on AndroidWorld~\cite{rawlesandroidworld}. 
\ours{} preserves the original task definitions, Android emulator interface, action space, and programmatic success verifier, while inserting the perturbation wrapper described in Section~\ref{sec:mdp} and illustrated in Figure~\ref{fig:benchmark_framework}. 
Each task is paired with one fixed perturbation instance, and the same task-perturbation pairs are shared across all models for reproducible and model-agnostic comparison. 
The perturbations cover \Statecolor{state}, \Actioncolor{action}, and \Transitioncolor{transition} components of the agent-environment loop; full perturbation configurations are provided in Appendix~\ref{app:experiment} and Table~\ref{tab:perturbation-params}.

\noindent \textbf{Evaluated models.}
We evaluate eight open-source mobile agents from GUI-OWL~\cite{ye2025mobileagentv3}, GUI-OWL-1.5~\cite{xu2026mobileagentv35}, UI-TARS~\cite{qin2025uitars}, and UI-TARS-1.5~\cite{qin2025uitars}, covering model scales from $2$B to $32$B. 
All models use their released prompts and observe the latest screenshot together with the action history. Due to the page limit, implementation, model serving, and logging details are reported in Appendix~\ref{app:experiment}.


\subsection{Robustness Evaluation}

We evaluate eight open-source mobile agents under both the clean and perturbed settings of \ours{}. 
Beyond aggregate task success rates, our MDP-grounded taxonomy enables us to break down robustness performance by perturbation class, revealing distinct failure profiles across the three components of the MDP interaction loop. 
Table~\ref{tab:evaluated_models} reports the per-class results, while the detailed categorical perturbation analysis (across full types) is shown in Figure~\ref{fig:GUIOwl15_result} and Figure~\ref{fig:GUIOwl8_result}.

\noindent ~\textbf{$\text{Observation}$: All agents degrade under real-world perturbation.}
Every evaluated agent suffers substantial degradation under perturbation, with absolute drops of $13$--$36$ points on overall success rate, uniformly across model families, parameter scales, and training paradigms. 
Performance loss spans all three MDP components rather than concentrating on a single class, indicating that current mobile agents lack the robustness needed for reliable deployment despite their strong canonical-benchmark performance. 
Notably, drops are comparable across very different agents, such as GUI-Owl-32B and GUI-Owl-1.5-8B, which both lose roughly $15$--$20$ points, suggesting that scaling and additional post-training alone do not close the robustness gap. The top-three challenging types are in Table~\ref{tab:most_influential_perturbation_types}.

\noindent $\bullet$~\textbf{State perturbations affect all agents most uniformly.}
Among the three perturbation classes, \Statecolor{\textbf{state-level}} perturbations cause the largest and most uniform degradation across agents. 
By modifying the rendered observation through font scale, UI theme, or orientation, these perturbations push UI elements outside the spatial regions that agents have learned to associate with their target functions, and agents fail to re-ground their coordinate predictions to the new layout. 
As a result, agents often issue coordinates aligned with the unperturbed layout, miss the intended target, and repeatedly retry the same erroneous coordinate without re-grounding against the current observation.

\noindent $\bullet$~\textbf{Action perturbations have the smallest effect.}
\Actioncolor{\textbf{Action-level}} perturbations, in contrast, lead to the smallest drops across most agents. 
This relative resilience comes from a property that current agents already possess by default: when an action does not produce the expected screen change because it was delayed, dropped, or absorbed by a frozen state, the agents naturally retry the same action on the next step. This implicit recovery behavior masks much of the underlying execution noise, particularly for short-lived perturbations such as \texttt{action\_delay} or \texttt{action\_drop}. 
Action-level robustness is therefore better characterized as a side-effect of the agents' default retry behavior.

\noindent $\bullet$~\textbf{Transition perturbations show the largest cross-agent variance.}
\Transitioncolor{\textbf{Transition-level}} perturbations exhibit the most variable behavior across agents: drops range from $-2.27$ on GUI-Owl-1.5-4B to $-29.54$ on GUI-Owl-1.5-2B and $-26.14$ on GUI-Owl-32B. 
This variance reflects qualitatively different agent strategies for handling unexpected intermediate states. Some agents recognize pop-ups, consent sheets, or notifications as off-task and dismiss them cleanly with only a small step-budget cost. Others treat the interruption as part of the task and engage deeply with multi-step dialogs, as shown in Figure~\ref{fig:caseMultisteps}, consuming most of their budget before reaching the actual goal.
The unusually small drop on GUI-Owl-1.5-4B is a notable exception, as its low clean-setting success rate ($43.10$) leaves limited headroom for further degradation, reminding us that absolute robustness drop must be interpreted alongside clean-setting capability.
We also observe a more subtle failure mode: even when an agent successfully dismisses a transition perturbation, a non-trivial fraction of subsequent attempts fail to complete the original task that the agent had solved correctly on the clean run, suggesting that interruptions disrupt the agent's goal memory and lead to wrong-subgoal resumption or premature termination. 
Transition perturbations, therefore, probe not only the agent's ability to handle off-trajectory states but also its capacity to maintain goal-directed memory across distractions.

\begin{table}[t!]
\centering
\small
\setlength{\tabcolsep}{2.2pt}
\renewcommand{\arraystretch}{1}
\caption{\small Evaluation results of open-source GUI-specialized agent models under clean and perturbed environments. Success rates are reported in percentages. Values in parentheses indicate absolute performance drops from the corresponding clean setting. $\Delta$Step reports the change in average step length from clean to perturbed settings, where red indicates increases and green indicates decreases. Numbers$^{\dagger}$ reported by the original paper~\cite{qin2025uitars}.}
\resizebox{\linewidth}{!}{
\begin{tabular}{lcccccccccccc}
\toprule
\textbf{Model} 
& \multicolumn{3}{c}{\textbf{Overall}} 
& \multicolumn{3}{c}{\textbf{State}} 
& \multicolumn{3}{c}{\textbf{Action}} 
& \multicolumn{3}{c}{\textbf{Transition}} \\
\cmidrule(lr){2-4}
\cmidrule(lr){5-7}
\cmidrule(lr){8-10}
\cmidrule(lr){11-13}
& 
\textbf{Clean} & \textbf{Pert.} & \textbf{$\Delta$Step}
& \textbf{Clean} & \textbf{Pert.} & \textbf{$\Delta$Step}
& \textbf{Clean} & \textbf{Pert.} & \textbf{$\Delta$Step}
& \textbf{Clean} & \textbf{Pert.} & \textbf{$\Delta$Step} \\
\midrule

GUI-Owl-7B 
& 63.79 & 48.28\textsuperscript{\dec{(-15.51)}} & \inc{+0.91}
& 63.64 & 40.91\textsuperscript{\dec{(-22.73)}} & \inc{+1.73}
& 60.71 & 48.21\textsuperscript{\dec{(-12.50)}} & \dec{-1.94}
& 65.91 & 55.68\textsuperscript{\dec{(-10.23)}} & \inc{+1.77} \\

GUI-Owl-32B 
& 64.66 & 41.81\textsuperscript{\dec{(-22.85)}} & \inc{+0.54}
& 65.91 & 47.73\textsuperscript{\dec{(-18.18)}} & \inc{+0.59}
& 60.71 & 35.71\textsuperscript{\dec{(-25.00)}} & \inc{+1.38}
& 65.91 & 39.77\textsuperscript{\dec{(-26.14)}} & \dec{-0.14} \\

\midrule

GUI-Owl-1.5-2B
& 48.71 & 26.29\textsuperscript{\dec{(-22.42)}} & \inc{+3.47}
& 50.00 & 27.27\textsuperscript{\dec{(-22.73)}} & \inc{+3.61}
& 41.07 & 30.36\textsuperscript{\dec{(-10.71)}} & \inc{+5.68}
& 52.27 & 22.73\textsuperscript{\dec{(-29.54)}} & \inc{+1.78} \\

GUI-Owl-1.5-4B
& 43.10 & 29.74\textsuperscript{\dec{(-13.36)}} & \inc{+1.72}
& 54.55 & 29.55\textsuperscript{\dec{(-25.00)}} & \inc{+4.02}
& 42.86 & 30.36\textsuperscript{\dec{(-12.50)}} & \dec{-0.10}
& 31.82 & 29.55\textsuperscript{\dec{(-2.27)}} & \dec{-0.25} \\

GUI-Owl-1.5-8B 
& 56.90 & 37.07\textsuperscript{\dec{(-19.83)}} & \inc{+3.77}
& 56.82 & 43.18\textsuperscript{\dec{(-13.64)}} & \inc{+7.12}
& 50.00 & 30.36\textsuperscript{\dec{(-19.64)}} & \dec{-0.46}
& 61.36 & 35.23\textsuperscript{\dec{(-26.13)}} & \inc{+2.29} \\

GUI-Owl-1.5-32B 
& 67.67 & 43.10\textsuperscript{\dec{(-24.57)}} & \inc{+3.54}
& 69.32 & 38.64\textsuperscript{\dec{(-30.68)}} & \inc{+7.12}
& 57.14 & 42.86\textsuperscript{\dec{(-14.28)}} & \inc{+1.21}
& 72.73 & 47.73\textsuperscript{\dec{(-25.00)}} & \inc{+1.05} \\

\midrule

UI-TARS-7B-SFT
& 33.00$^{\dagger}$ & 25.86\textsuperscript{\dec{(-7.14)}} & --
& -- & 29.55\textsuperscript{\dec{(-3.45)}} & --
& -- & 28.57\textsuperscript{\dec{(-4.43)}} & --
& -- & 20.45\textsuperscript{\dec{(-12.55)}} & -- \\

UI-TARS-1.5-7B 
& 64.20$^{\dagger}$ & 27.59\textsuperscript{\dec{(-36.61)}} & --
& -- & 29.55\textsuperscript{\dec{(-34.65)}} & --
& -- & 25.00\textsuperscript{\dec{(-39.20)}} & --
& -- & 27.27\textsuperscript{\dec{(-36.93)}} & -- \\

\bottomrule
\end{tabular}
}
\label{tab:evaluated_models}
\end{table}

\subsection{Error Analysis}
\label{sec:error_analysis}

\begin{figure}[h!]
    \centering
    \includegraphics[width=\linewidth]{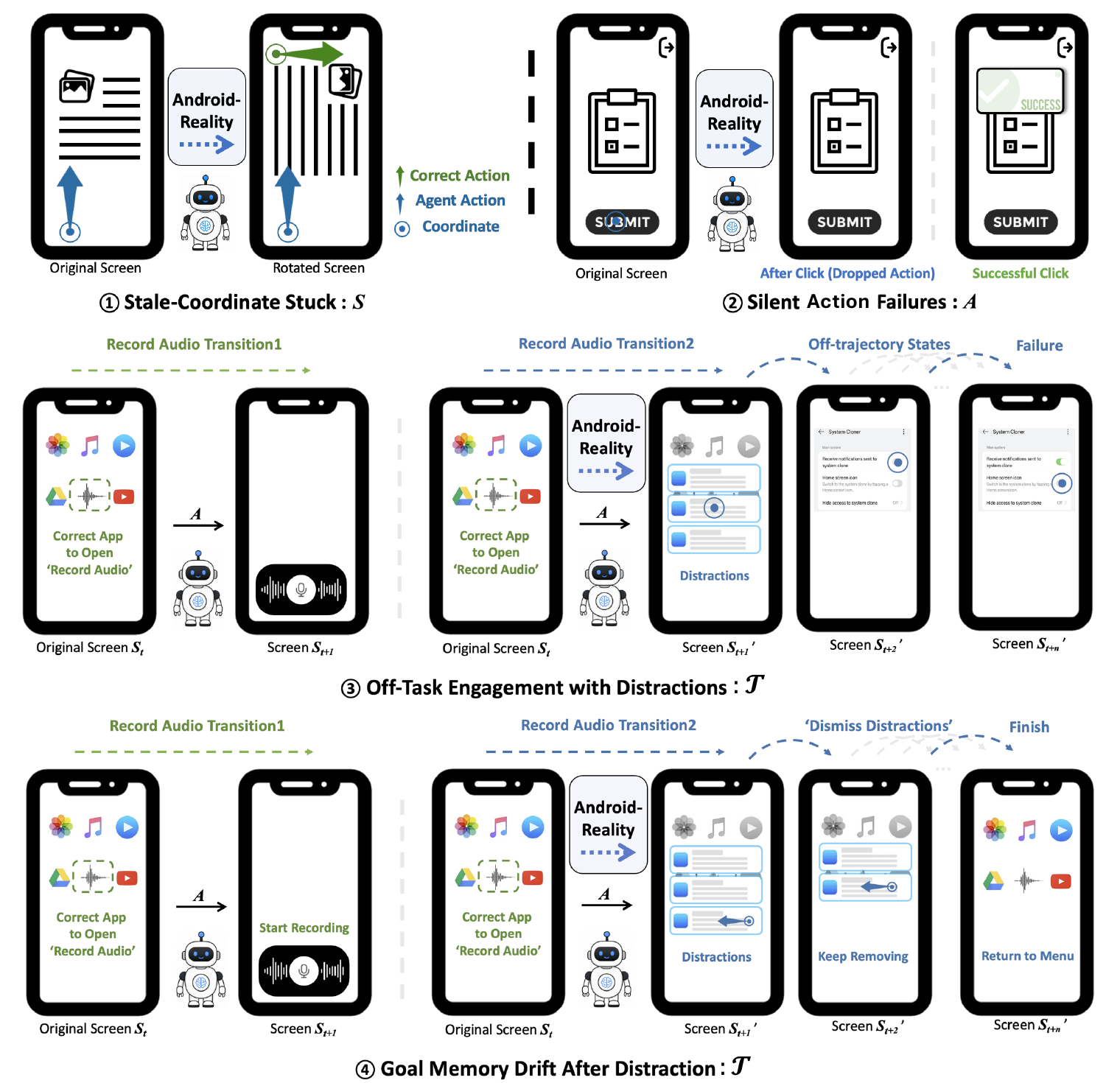}
    \caption{\textbf{Visual illustration of the four error categories surfaced by \ours{}.}
    (1) \textbf{Stale-Coordinate Stuck} ($\mathcal{S}$): under state-level perturbations such as orientation changes, the agent's predicted coordinate misses the rendered target, and the agent repeatedly retries the same incorrect coordinate. 
    (2) \textbf{Silent Action Failures} ($\mathcal{A}$): action-level perturbations silently drop the issued action; the agent does not notice the missed click and proceeds as if it had succeeded. 
    (3) \textbf{Off-Task Engagement with Distractions} ($\mathcal{T}$): a transition-level perturbation injects off-trajectory states (e.g., system pop-ups), and the agent treats them as part of the task, navigating deeper into the distraction until failure. 
    (4) \textbf{Goal Memory Drift After Distraction} ($\mathcal{T}$): the agent correctly dismisses the distraction but loses track of the original task, returning to a menu or terminating without completing the intended workflow. The symbols in parentheses denote the MDP component each error class is most strongly associated with, in correspondence to this work's benchmark design.}
    \label{fig:error_categories}
\end{figure}

To understand \emph{why} agents fail under perturbation, we conducted a systematic error analysis over all failed trajectories from our main evaluation. 
Failures were tagged by manually inspecting per-step screenshots, action logs, and the agent's reasoning traces. 
We identify four recurring error categories that together account for most of the perturbation-induced failures across all evaluated models.
Figure~\ref{fig:error_categories} provides a schematic illustration of the four categories of recurring failures.

\textbf{\ding{172} Stale-Coordinate Stuck.}
The most striking failure pattern occurs under \Statecolor{state-level} perturbations. When display density, theme, or orientation alters the rendered layout, the agent often emits coordinates that miss the new position of the target element and land on empty space or an adjacent UI element. 
After observing that the screen has not changed in the expected way, the agent does not re-ground its prediction to the new layout; instead, it retries the same incorrect coordinate-level actions repeatedly, in extreme cases up to dozens of identical clicks before the step budget is exhausted. 
This pattern is observed uniformly across all four model sizes, suggesting that current agents lack an explicit re-grounding behavior when an action fails to produce the expected state change. 

\noindent \textbf{\ding{173}  Silent Action Failures.} Under \Actioncolor{action-level} perturbations such as \texttt{action\_drop\_random}, the action generated by the agent is silently delayed, dropped, or absorbed by an unresponsive screen. Although agents naturally retry the same action on the next step (which masks much of the underlying noise on simpler tasks), a non-trivial fraction of failures occur when the agent does \textbf{\emph{not}} register that the previous action had no effect, and proceeds with subsequent steps in its plan as if it had succeeded. 
The result would be a trajectory that is internally coherent but desynchronized from the actual environment state, and that often ends with the agent declaring task completion on a screen that never advanced. 
This failure mode is a direct consequence of agents lacking an explicit verification step between action generation and continuation. There is an example shown in Figure~\ref{fig:error_categories} (\ding{173}).

\noindent \textbf{\ding{174}}  \textbf{Off-Task Engagement with Distractions.}
Under \Transitioncolor{transition-level} perturbations such as pop-ups, consent sheets, advertisements, and notifications, the agent often fails to recognize the interruption as off-task and instead treats it as part of the intended workflow. 
This manifests in several ways: the agent may engage deeply with multi-step dialogs, navigate through fake interaction flows until the step budget is exhausted, or treat the interruption's dismissal as task completion and emit \texttt{terminate(success)} without ever attempting the original task. 
Across these variants, the underlying issue is the same: the agent does not distinguish between off-trajectory states introduced by the environment and on-trajectory states relevant to the goal. 
The original task may never even begin once the distraction takes hold of the agent's attention. The illustration is shown in Figure~\ref{fig:error_categories} (\ding{174}).

\noindent \textbf{\ding{175}}  \textbf{Goal Memory Drift After Distraction.}
A more subtle and consequential failure mode arises under \Transitioncolor{transition-level} perturbations: the agent \emph{successfully} dismisses an interruption, yet subsequently fails to complete the original task. 
In a non-trivial fraction of trajectories, the agent correctly identifies the pop-up or notification as off-task, dismisses it cleanly, and then either resumes on the wrong subgoal, restarts the task from scratch, or terminates prematurely as if the task were already complete. 
These are tasks the same agent solves correctly on the clean run, indicating that the failure is not caused by an inability to handle the interruption itself, but by the disruption it imposes on the agent's working memory of the original task plan. 
This category reveals a robustness dimension that aggregate success rates conflate with raw distraction-handling failures: even when the dismissal is correct, the cognitive cost of the interruption can derail the trajectory.  Similarly, please refer to Figure~\ref{fig:error_categories} (\ding{175}) for illustration.



\section{Test-Time Introspective Recovery}
\begin{wrapfigure}{r}{0.45\linewidth}
    \centering
    \includegraphics[width=\linewidth]{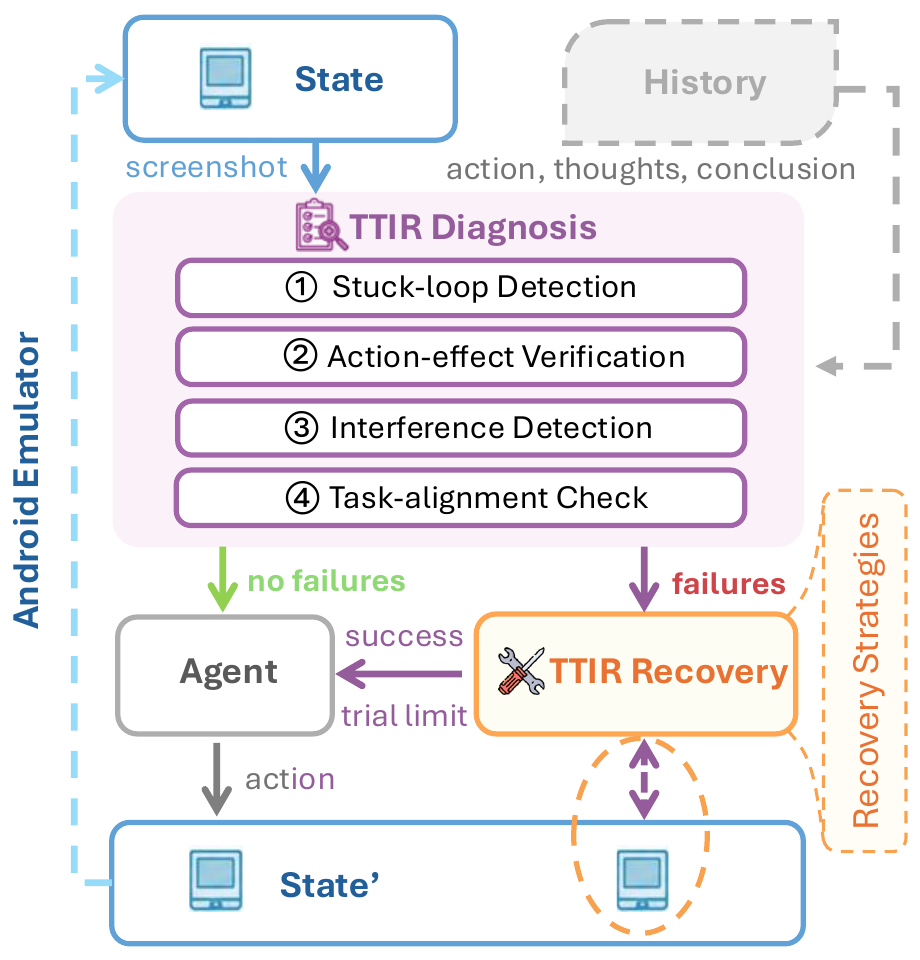}
    \caption{\small The TTIR framework. TTIR diagnoses four failure signals from our error analysis: stuck-loop behavior, unsatisfied action effects, off-task interference, and task misalignment. When a failure is detected, it invokes a limited recovery session before returning the recovered state to the agent--emulator loop.}
    \label{fig:ttir_process}
    \vspace{-15mm}
\end{wrapfigure}





The error analysis in Section~\ref{sec:error_analysis} surfaced four failure categories that share a common root cause: the agent does not check whether its current state remains consistent with successful task progress before continuing. 
This observation suggests that robustness can be improved by introducing an explicit self-monitoring mechanism at test time, without retraining the underlying agent. 
We propose \textbf{Test-Time Introspective Recovery (TTIR)}: a training-free, model-agnostic mechanism that augments the agent's interaction loop with a \emph{diagnose-then-recover} step before each action. 
At every step, a lightweight diagnoser inspects the current screenshot together with the recent action history and asks four targeted questions, each designed to detect one of the failure categories identified in our analysis. 
When a problem is detected, TTIR dispatches a recovery strategy specifically tailored to that category. 
The diagnoser and recovery agent both reuse the same backbone VLM as the main agent, introducing no additional parameters or training data.

\subsection{TTIR Module Design}
\label{sec:ttir_design}


TTIR is organized around a one-to-one correspondence between the four error categories from the Section~\ref{sec:error_analysis} and four `diagnose-then-recover strategies' (dashed orange box on the right side of Figure~\ref{fig:ttir_process}). 
At each step, the `TTIR Diagnosis Module' (i.e., diagnoser) produces a structured JSON judgment of the current state along four dimensions, each derived from a targeted question. 
If any dimension flags a problem, TTIR enters a bounded recovery session in which the recovery agent is invoked through a recovery-specific prompt that declares the diagnosed problem, supplies the recovery target condition, and specifies the relevant constraints. 
The recovery agent then generates several corrective actions tailored to that specific failure pattern, and control returns to the main agent once the diagnosed problem is resolved or the recovery budget is exhausted. 
TTIR therefore provides a pure test-time improvement to the agent's interaction loop.
Figure~\ref{fig:ttir_process} illustrates the overall design of TTIR. 
The four strategies are summarized as follows: we provide details on the diagnostic signals and recovery prompts for each strategy in Appendix~\ref{app:ttir_routines}.

\textbf{{\ding{172} Stuck-loop detection $\Rightarrow$ alternative-action recovery.}}
TTIR detects stuck loops using both model-based and deterministic signals. The diagnoser checks whether recent actions form a repeated unresolved attempt and outputs \texttt{repeat\_failure\_detected}. TTIR also triggers this diagnosis when the last $N$ steps (default $N{=}3$) have identical action, summary, and thought. Once triggered, TTIR applies an \emph{alternative-action} recovery: the recovery agent is told that the previous action has repeatedly failed and is instructed to choose a different visible target or route toward the same sub-goal instead of retrying the same coordinate. 

\textbf{\ding{173} Action-effect verification $\Rightarrow$ alternative-action recovery.}
The `TTIR Diagnosis' module compares the agent's previous intended effect against the current screenshot, and it will flag the cases where the action was executed, while its expected outcome is not visible. After that, the `TTIR Recovery' module was prompted to try another way to achieve the same intended effect.



\begin{wrapfigure}{r}{0.45\linewidth}
    \centering
    \includegraphics[width=\linewidth]{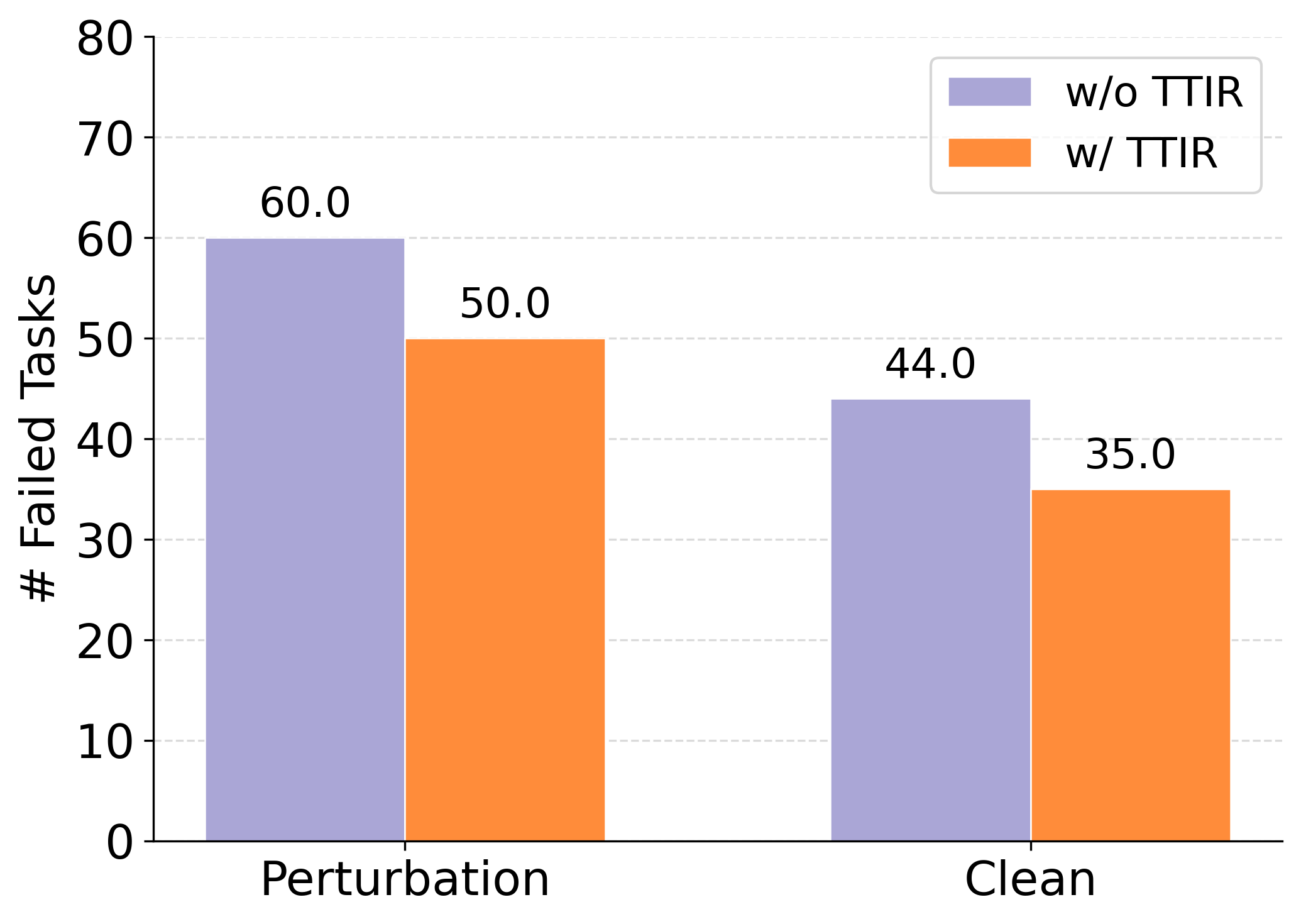}
    \caption{\small Comparison of failed task counts with and without TTIR (Ours). It shows a notable reduction in failures under the Perturbation condition, while also rectifying erroneous task outcomes in the Clean condition.}
    \label{fig:ttir_results}
    \vspace{-15mm}
\end{wrapfigure}

\textbf{\ding{174} Interference detection $\Rightarrow$ dismiss-interference recovery.}
The `TTIR Diagnosis' module inspects the screenshot for off-task overlays such as pop-ups, consent sheets, advertisements, and notifications. 
Then, the `TTIR Recovery' was prompted to close, decline, or navigate past the interfering element, while explicitly forbidding engagement with the interruption's content.

\textbf{\ding{175} Task-alignment check $\Rightarrow$ re-orientation recovery.}
The diagnoser checks whether the current screen still belongs to the original task path, particularly after a perturbation has been handled. 
The recovery agent was prompted to find an action that restores the correct task context (e.g., navigating back or re-opening the relevant app), without attempting to complete the task itself.

\subsection{Empirical Study}

To evaluate the effectiveness of TTIR, we apply it to GUI-Owl-7B and compare its performance with and without TTIR on both the perturbed and clean settings. 
TTIR is applied as a pure test-time setting, with no fine-tuning or reinforcement learning. 
Figure~\ref{fig:ttir_results} reports the number of failed tasks before and after applying TTIR.

In the \textbf{perturbed setting}, TTIR reduces the number of failed tasks from 60 to 50 (a 17\% relative reduction). 
More notably, in the \textbf{clean setting}, TTIR still reduces failed tasks from 44 to 35 (a 20\% relative reduction), even though no perturbation is injected. 
This is consistent with our observation in Section~\ref{sec:error_analysis} that the four error categories are not artifacts of perturbation but latent weaknesses already present in current mobile agents, and indicates that the gains stem from genuine improvements in agent introspection rather than perturbation-specific countermeasures. 
We provide one qualitative case study for each of the four recovery strategies in Appendix~\ref{app:recovery_cases}.

\section{Conclusion}
\vspace{-3mm}
In this work, we investigated robustness as a dimension of mobile agent evaluation that is currently underexplored. 
We formalized real-world interface variability through a Markov Decision Process (MDP) perspective, deriving a perturbation taxonomy organized by which MDP component each perturbation acts upon.
Based on this taxonomy, we constructed \textbf{\ours{}}, it is an online mobile benchmark built on AndroidWorld that injects realistic and controllable perturbations along state, transition, and action axes, which can support fine-grained robustness evaluation across mobile agents. 
Our evaluation of eight mobile agents on \ours{} consistently showed substantial performance degradation under categorized perturbation, and we conducted a comprehensive systematic error analysis that surfaced four recurring failure categories. Motivated by this analysis, we further proposed \text{Test-Time Introspective Recovery (TTIR)}, a training-free mechanism that pairs each error category with a targeted diagnose-and-recover routine, mitigating failures on both perturbed and clean settings.
We expect the \ours{} can serve as a useful framework for the research community to study the robustness in mobile-agent evaluation and to push for real-world use by surfacing failure modes that are otherwise difficult to observe under clean benchmark conditions.

\clearpage


\bibliographystyle{unsrtnat}
\bibliography{ref}

\newpage
\appendix

\appendix





\section{\ours{} Specifications and Visualization}
\label{app:framework_spec}


\begin{figure}[h!]
    \centering
    \includegraphics[width=\linewidth]{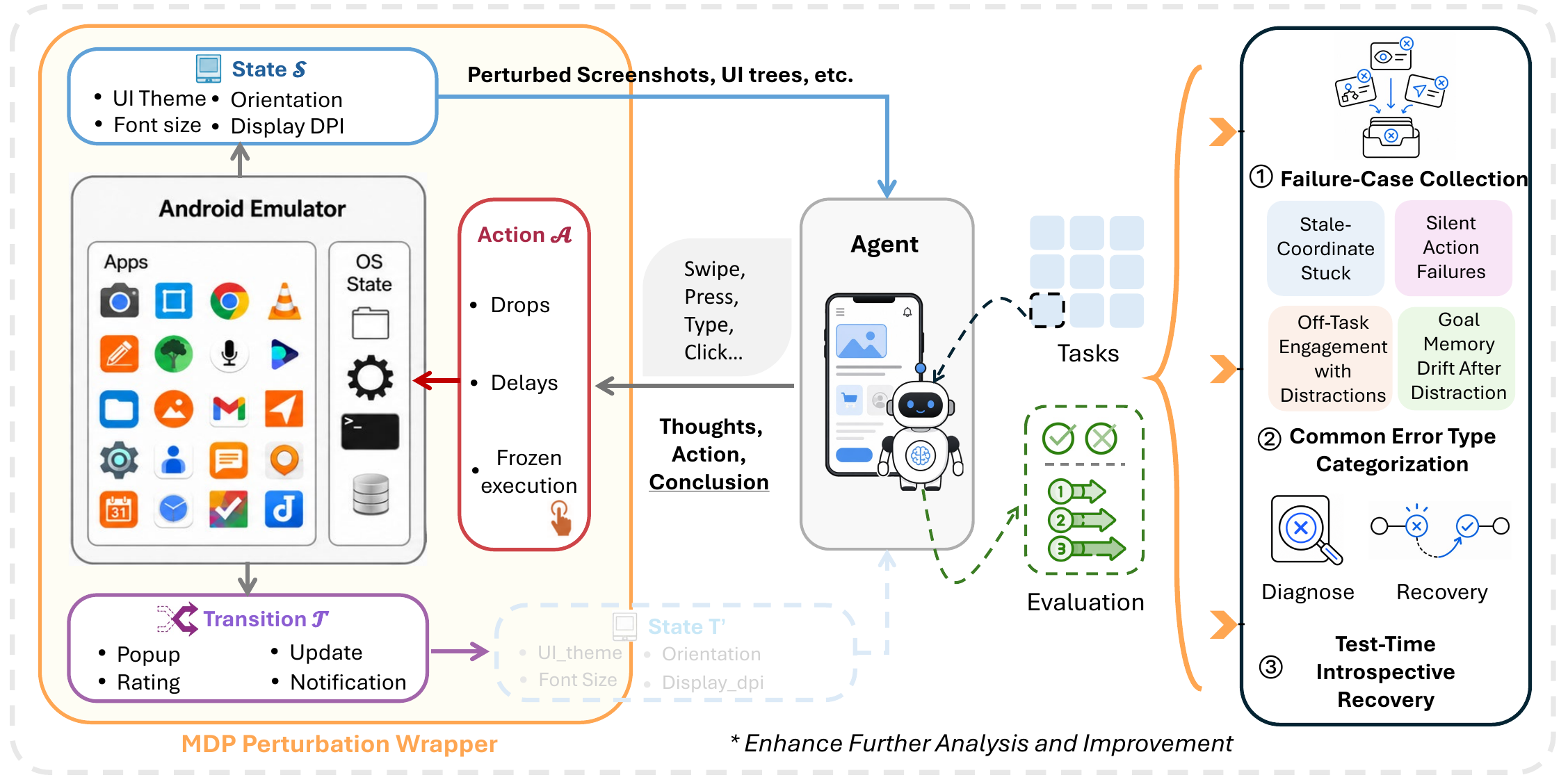}
    \caption{Overview of the proposed \ours{} benchmark. 
Starting from a clean AndroidWorld environment, we wrap the agent--environment interaction loop with an MDP perturbation layer that injects realistic and controllable variations along three axes: \Statecolor{\textbf{State-level}} perturbations that alter UI observations such as UI theme, orientation, font size and display dpi; \Transitioncolor{\textbf{Transition-level}} perturbations that introduce intermediate environment dynamics such as pop-ups, notifications, update, rating and notification; and \Actioncolor{\textbf{Action-level}} perturbations that corrupt the mapping from intended actions to executed actions through drops, delays, or frozen. 
The agent receives perturbed screenshots, UI trees, and APIs, then performs thoughts, actions, and conclusions over the task set. 
This design converts a single canonical benchmark MDP into a family of perturbed MDPs that preserve task semantics while exposing robustness failures under real-world interface variability. Beyond robustness evaluation, the benchmark supports analysis and improvement, such as failure cases collection, which is then grouped into common error types, and finally used to guide test-time introspective recovery.}
    \label{fig:benchmark_framework}
\end{figure}

\subsection{Perturbation Taxonomy}

In this section, we enumerate all 24 perturbations in \ours{}, listing their MDP class, name, description, and tunable parameters in Table~\ref{tab:perturbation_taxonomy}.

\begin{table*}[t!]
\centering
\small
\setlength{\tabcolsep}{4pt}
\renewcommand{\arraystretch}{1.0}
\caption{Perturbation taxonomy in \ours{}. Each perturbation preserves the original task goal but modifies a specific MDP component of the `agent and environment' interaction loop. All perturbations are \textbf{fully parameterized} along three orthogonal axes, which are: \emph{magnitude}, \emph{timing}, and \emph{persistence}, and it enables controllable severity sweeps and reproducible evaluation. Tunable parameters that determine perturbation \textbf{strength} are highlighted in \textbf{bold}.}
\label{tab:perturbation_taxonomy}

\begin{adjustbox}{max width=\textwidth, max totalheight=0.9\textheight, keepaspectratio}
\begin{tblr}{
  colspec = {
    Q[c,m,0.10\linewidth]
    Q[l,m,0.21\linewidth]
    Q[l,m,0.34\linewidth]
    Q[l,m,0.30\linewidth]
  },
  rowsep = 1pt,
  row{1} = {bg=backheader},
  cell{2}{1}  = {r=7}{c,m},
  cell{9}{1}  = {r=6}{c,m},
  cell{15}{1} = {r=10}{c,m},
  cell{15}{4} = {r=10}{l,m},
  row{2,4,6,8} = {bg=backblue},
  row{9,11,13} = {bg=backred},
  row{15,17,19,21,23} = {bg=backpurple},
  hline{1,2,9,15,25} = {-}{},
}
\textbf{MDP Comp.} & \textbf{Perturbation} & \textbf{Description} & \textbf{Tunable Parameters} \\

\Statecolor{\textbf{State}}
& \pert{display_size_dpi_compact}
& Compresses display density to test robustness to compact screen configurations.
& \textbf{Density scale}, \textbf{size scale}, trigger step, settle time, restore-on-close \\

& \pert{display_size_dpi_zoomed_out}
& Alters display scale and DPI to test recognition under altered visual sizing.
& \textbf{Density scale}, \textbf{size scale}, trigger step, settle time, restore-on-close \\

& \pert{font_size_large} / \pert{small}
& Changes text rendering, potentially shifting layout and element positions.
& \textbf{Font scale}, trigger step, settle time, restore-on-close \\

& \pert{ui_theme_dark}
& Switches the interface to dark mode, altering colors and contrast.
& \textbf{Theme mode}, trigger step, settle time, restore-on-close \\

& \pert{orientation_landscape}
& Rotates the display to landscape, changing spatial layout.
& \textbf{Orientation sequence}, trigger step, settle time, restore-on-close \\

& \pert{locale_date_format}
& Switches locale-dependent date and time formats (US12 / EU24).
& \textbf{Date format}, \textbf{time format}, trigger step, settle time, restore-on-close \\

& \pert{partial_loading_skeleton}
& Renders skeleton placeholders to simulate partially loaded screens.
& \textbf{Duration steps}, trigger step, block actions, content text \\

\Actioncolor{\textbf{Action}}
& \pert{action_delay_realistic}
& Delays action execution to simulate latency in real mobile interactions.
& \textbf{Delay probability}, \textbf{delay seconds}, random seed \\

& \pert{action_drop_random}
& Randomly drops issued actions, testing detection of failed operations.
& \textbf{Drop probability}, random seed \\

& \pert{execution_delay_loading}
& Introduces post-action loading delays, requiring waiting vs.\ failure judgment.
& \textbf{Duration steps}, trigger step, block actions, loading text \\

& \pert{execution_delay_frozen}
& Freezes execution after an action, simulating temporary unresponsiveness.
& \textbf{Duration steps}, trigger step, block actions, frozen-screen text \\

& \pert{app_hang_restart_required}
& Simulates an application hang requiring restart or recovery.
& \textbf{Trigger step}, dialog text, settle time, recovery options \\

& \pert{state_reset_home}
& Resets the UI to home, testing recovery after losing task context.
& \textbf{Trigger step}, return-home flag, settle time \\

\Transitioncolor{\textbf{Transition}}
& \pert{popup_permission}
& Permission dialogs that alter the immediate action context.
& \makecell[l]{
  \textbf{Trigger step}\\
  \textbf{Surface style}\\
  \textbf{Blocking}\\
  \textbf{Auto-dismiss duration}\\
  Content title\\
  Content message\\
  Content buttons\\
  Primary/secondary actions\\
  Profile pool
} \\

& \pert{popup_security}
& Security-related pop-ups that test interruption handling.
&  \\

& \pert{bottom_sheet_consent}
& Consent-style bottom sheets requiring dismissal or decision making.
&  \\

& \pert{interstitial_ad}
& Full-screen ad interruptions that block the original trajectory.
&  \\

& \pert{ad_trial_offer}
& Trial-offer advertisements as unexpected intermediate states.
&  \\

& \pert{rate_dialog}
& Rating prompts, a common non-task dialog in mobile apps.
&  \\

& \pert{notification_message}
& Message notifications that may distract the agent from its task.
&  \\

& \pert{notification_delivery}
& Notification-style interruptions related to delivery events.
&  \\

& \pert{update_app_sheet}
& App-update prompts that redirect the interaction trajectory.
&  \\

& \pert{update_system}
& System-update prompts emulating OS-level interruptions.
&  \\

\end{tblr}
\end{adjustbox}

\end{table*}

\subsection{Experimental Parameter Configurations}

For reproducibility, Table~\ref{tab:perturbation-params} reports the default parameter values used for each tunable perturbation in our main experiments. 
All evaluations reported in the main text use these default values unless otherwise specified.

\begin{table}[t]
\centering
\small
\caption{Default parameter values used for the tunable perturbations in our main experiments. These parameters jointly control each perturbation and remain fixed across all evaluated agents to ensure reproducible and model-agnostic comparison. }
\label{tab:perturbation-params}
\begin{tabular}{l l c}
\toprule
\textbf{Type} & \textbf{Parameter} & \textbf{Default Value} \\
\midrule
Font scaling & \texttt{FONT\_SIZE\_LARGE\_SCALE} & 1.35 \\
Font scaling & \texttt{FONT\_SIZE\_SMALL\_SCALE} & 0.82 \\
Display scaling & \texttt{DISPLAY\_COMPACT\_DENSITY\_SCALE} & 0.85 \\
Display scaling & \texttt{DISPLAY\_COMPACT\_SIZE\_SCALE} & 0.94 \\
Display scaling & \texttt{DISPLAY\_ZOOMED\_OUT\_DENSITY\_SCALE} & 0.76 \\
Display scaling & \texttt{DISPLAY\_ZOOMED\_OUT\_SIZE\_SCALE} & 0.88 \\
Orientation & \texttt{ORIENTATION\_PROFILE\_SEQUENCE} & \texttt{landscape} \\
App responsiveness & \texttt{APP\_HANG\_TRIGGER\_STEPS} & 2 \\
Action delay & \texttt{ACTION\_DELAY\_PROBABILITY} & 0.15 \\
Action delay & \texttt{ACTION\_DELAY\_SECONDS} & 1.5 \\
Action delay & \texttt{ACTION\_DELAY\_SEED} & 0 \\
Action drop & \texttt{ACTION\_DROP\_PROBABILITY} & 0.08 \\
Action drop & \texttt{ACTION\_DROP\_SEED} & 0 \\
\bottomrule
\end{tabular}
\end{table}

\section{Experimental Details}
\label{app:experiment}

This appendix expands the experimental setup summarized in Section~\ref{sec:setup}.

\paragraph{Benchmark construction.}
We conduct experiments on \ours{}, which is built on top of AndroidWorld~\cite{rawlesandroidworld}. 
The benchmark keeps the original AndroidWorld task definitions, Android emulator interface, action space, and programmatic success verifier unchanged. 
The only modification is the insertion of the perturbation wrapper from Section~\ref{sec:mdp}, which intercepts the agent--environment loop and applies controlled perturbations to the state, action, or transition component of the MDP. 
Figure~\ref{fig:benchmark_framework} illustrates the overall framework of \ours{}, including the perturbation wrapper, the AndroidWorld task environment, and the agent-emulator interaction loop.
Thus, the task goal and reward definition remain fixed, while the interaction conditions are made more realistic.

\paragraph{Perturbation instantiation.}
Each task is paired with exactly one perturbation instance. 
The perturbation random seed is fixed, so the same task--perturbation pair is realized identically for all evaluated models. 
Unless otherwise specified, each perturbation uses a fixed default severity. 
\Statecolor{State-level} perturbations persist throughout the episode as global interface configurations, including font size, display density, UI theme, orientation, locale/date format, and partial-loading skeleton screens. 
\Actioncolor{Action-level} perturbations modify the execution of intended actions under seed-controlled probabilities, including action delays, dropped actions, loading or frozen states, app hangs, and home resets. 
\Transitioncolor{Transition-level} perturbations are injected at task-specific runtime steps and introduce unexpected intermediate states, including advertisements, consent sheets, permission dialogs, security pop-ups, rating prompts, notifications, and app/system update prompts. 
The full parameter list and default values are provided in Table~\ref{tab:perturbation-params}; the evaluation protocol is described in Section~\ref{sec:eva}.


\paragraph{Evaluated models.}
We evaluate eight open-source mobile agents from GUI-OWL~\cite{ye2025mobileagentv3}, GUI-OWL-1.5~\cite{xu2026mobileagentv35}, UI-TARS~\cite{qin2025uitars}, and UI-TARS-1.5~\cite{qin2025uitars}. 
The selected models cover $2$B to $32$B parameter scales and include both supervised fine-tuning and reinforcement-learning-based training paradigms. 
This model suite allows us to test whether robustness failures are model-specific or shared across current mobile-agent systems.

\paragraph{Model serving.}
For all experiments, agents observe the most recent screenshot and their action history. 
We use the original released prompts for each model. 
All models are served with vLLM using sampling temperature $0.1$. 
Models with $\leq 8$B parameters are served on a single NVIDIA A100 GPU, while $32$B models are served on two NVIDIA A100 GPUs. 
The Android emulator runs on a separate CPU host and communicates with the inference server through the perturbation wrapper.

\paragraph{Execution protocol.}
For each model, we evaluate the full task set under both clean and perturbed settings. 
The clean setting is the original AndroidWorld environment, and the perturbed setting is \ours{} with one perturbation profile applied to each task. 
Each task--perturbation pair is evaluated once under the fixed seed. 
Because perturbations may introduce additional intermediate states, interruptions, or delayed responses, all clean and perturbed runs use a step budget of $20\times$ the human reference length.

\paragraph{Logging.}
During evaluation, we record the full interaction trajectory for each run, including the task goal, perturbation configuration, per-step screenshots, action history, issued actions, model reasoning traces, and final verification result. 
For both clean AndroidWorld and perturbed \ours{} settings, each run uses a step budget of $20\times$ the human reference length to account for perturbation-induced intermediate states, delayed responses, and recovery attempts. 
For failed runs, we retain the complete trajectory to support the error analysis in Section~\ref{sec:error_analysis} and the TTIR intervention study in Section~\ref{sec:ttir_design}. 
These logs enable manual attribution of failures to visual grounding errors, silent execution failures, off-task engagement with distractions, or goal-memory drift after interruptions.

\section{Error Category Summary}

This section provides a detailed reference for the four error categories identified in our analysis (Section~\ref{sec:error_analysis}). 
We first present a visual illustration of all four categories, followed by one
representative example drawn from real failure trajectories.

\subsection{Representative Failure Example}

\begin{figure}[h!]
    \centering
    \includegraphics[width=0.99\linewidth]{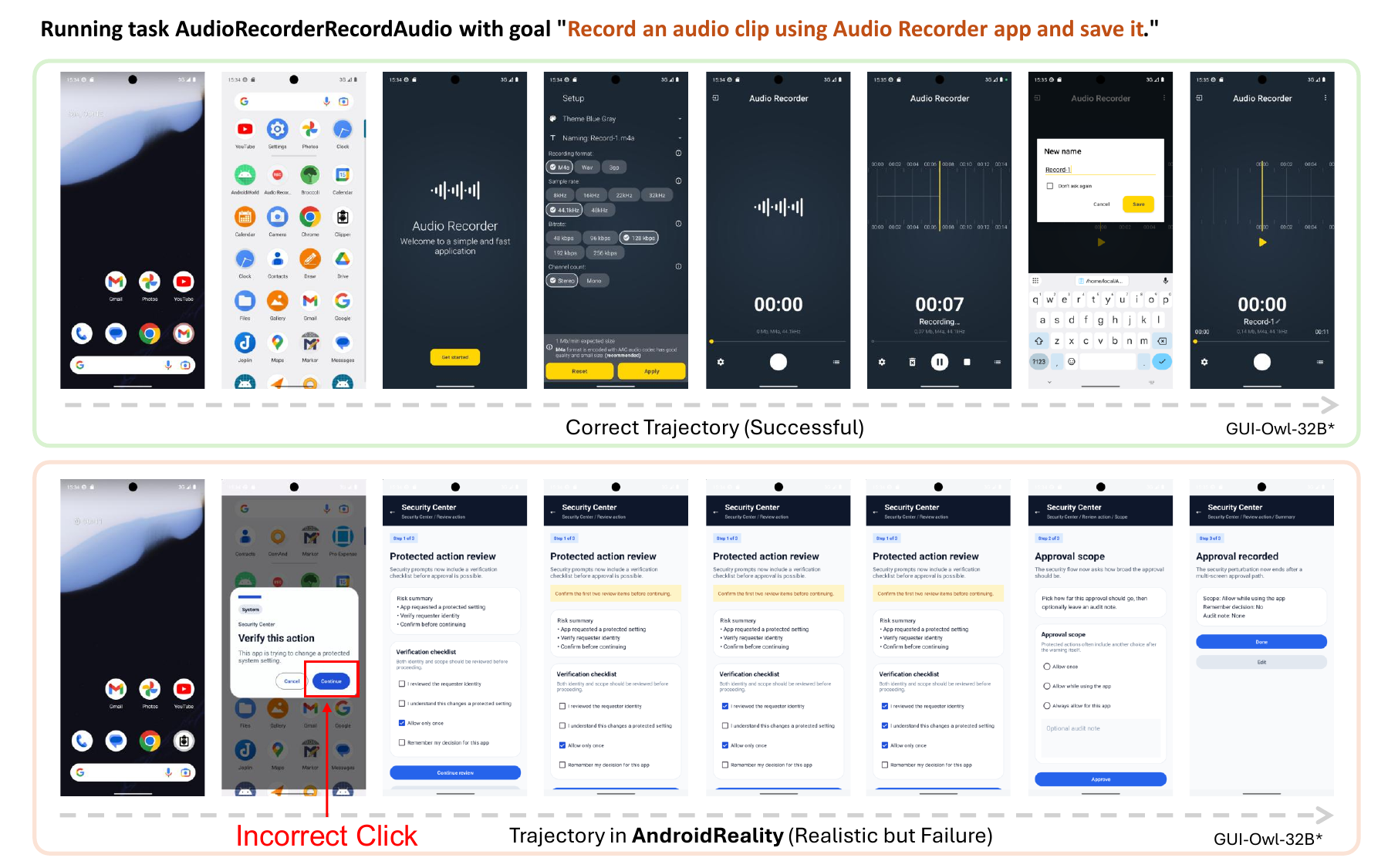}
    \caption{The actual failure case analysis of GUI-Owl-32B under a clean environment and AndroidReality. In the clean setting, the model successfully completes the task by opening the Audio Recorder app, recording an audio clip, and saving it with the correct name. In AndroidReality, however, a realistic security pop-up interrupts the interaction flow. The model fails to properly handle this unexpected intermediate state: it clicks into the security center and gets trapped in the security settings, and forgets the actual task to accomplish, eventually leading to task failure. This example illustrates how AndroidReality exposes robustness challenges that are hidden in clean environments, especially the difficulty of recovering from realistic mobile interruptions.}
    \label{fig:caseMultisteps}
\end{figure}

To complement the schematic illustration in Figure~\ref{fig:error_categories}, we present a real failure trajectory from \ours{} that demonstrates how a perturbation can derail an otherwise solvable task. 
Figure~\ref{fig:caseMultisteps} shows a representative case in which the agent encounters a transition-level perturbation in the form of an unexpected pop-up. 
Although the agent initially follows the correct task plan, the appearance of the pop-up diverts its attention away from the original goal: the agent engages with the pop-up content, follows its prompts, and never resumes the intended workflow. 
The trajectory ultimately ends in failure, despite the same agent successfully completing this task in the clean setting. 
This example concretely illustrates the failure mode introduced in Section~\ref{sec:error_analysis} and shows how a single transition-level perturbation can trigger task failure through distraction-induced trajectory drift.

\begin{figure}
    \centering
    \includegraphics[width=\linewidth]{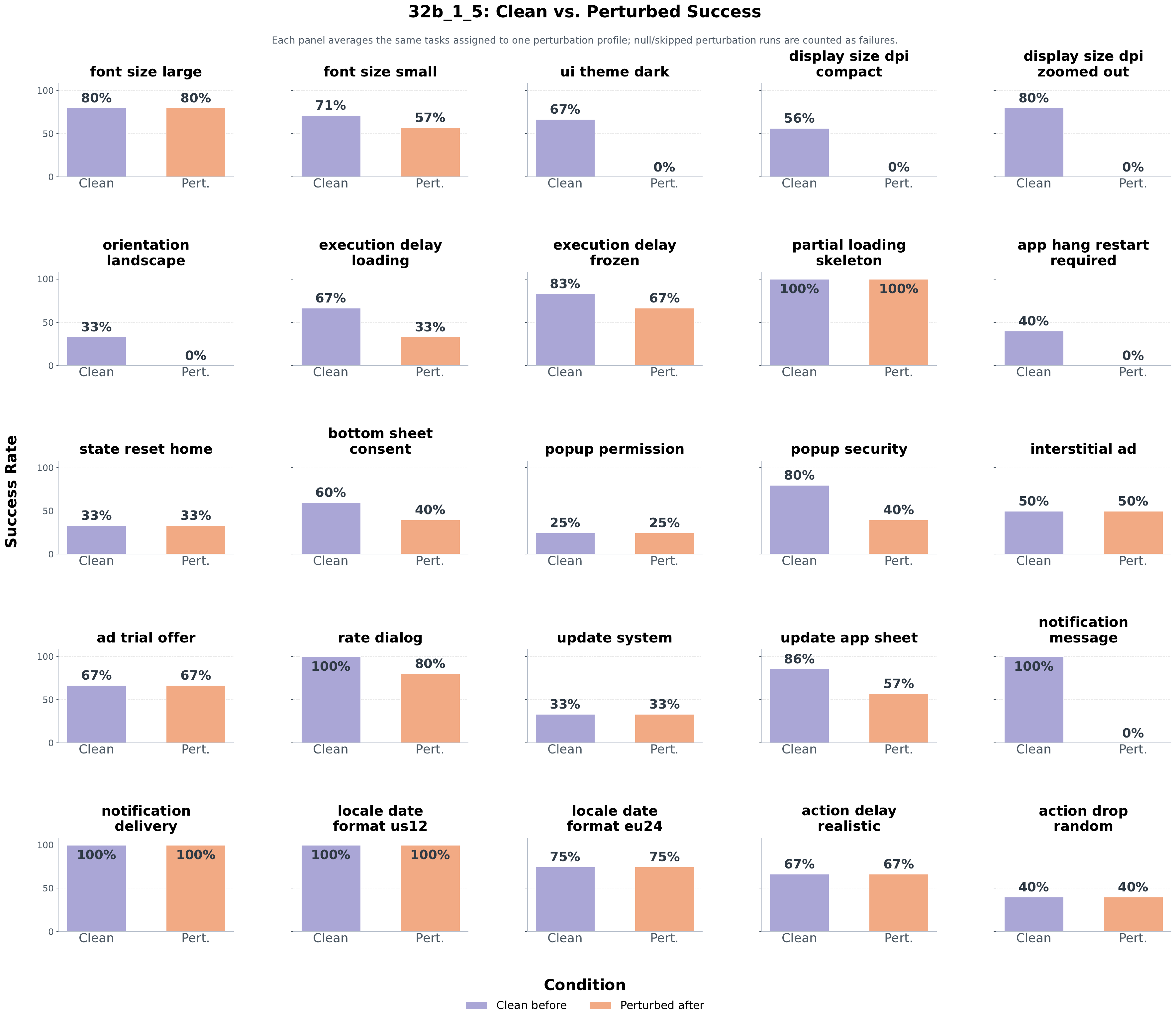}
    \caption{`Clean-before' (before applying AndroidReality) vs.\  `Perturbed-after' (After applying all reality perturbation profiles) success rates of GUI-Owl-1.5-32B across individual AndroidReality perturbation profiles. 
Each subplot reports the success rate on the same task subset before and after applying one perturbation type, where null or skipped perturbed runs are counted as failures. 
The results show that robustness degradation is highly perturbation-specific. 
State-level perturbations that substantially alter visual grounding, such as UI theme changes, display-density/size changes, and landscape orientation, cause the sharpest drops, in several cases reducing success to zero. 
In contrast, format-preserving perturbations such as locale/date-format changes or partial-loading skeletons are less disruptive (note, it also varies on the model's ability). 
Action-level perturbations show mixed effects: some execution failures such as loading delays and app hangs reduce performance, whereas short-lived delays or dropped actions can be partially masked by the agent's default retry behavior. 
Transition-level perturbations also vary widely, with security pop-ups, update sheets, and notification messages exposing failures in distraction handling and task continuity. 
Overall, this fine-grained breakdown shows that AndroidReality reveals not only aggregate robustness degradation, but also which concrete real-world interface variations most strongly challenge current mobile agents.}
    \label{fig:GUIOwl15_result}
\end{figure}

\begin{table}[h!]
\centering
\small
\setlength{\tabcolsep}{3.2pt}
\renewcommand{\arraystretch}{1.05}
\caption{\small Most influential perturbation types, defined as perturbation profiles for which all six models show lower success rates after perturbation. Success rates are reported in percentages. $\Delta$ denotes the absolute change from clean to perturbed performance, i.e., Perturbed $-$ Clean.}
\resizebox{\linewidth}{!}{
\begin{tabular}{lcccccc}
\toprule
\textbf{Model}
& \multicolumn{2}{c}{\textbf{Notification Message}}
& \multicolumn{2}{c}{\textbf{Display Size DPI Compact}}
& \multicolumn{2}{c}{\textbf{Execution Delay Frozen}} \\
\cmidrule(lr){2-3}
\cmidrule(lr){4-5}
\cmidrule(lr){6-7}
& \textbf{Clean $\rightarrow$ Pert.} & \textbf{$\Delta$}
& \textbf{Clean $\rightarrow$ Pert.} & \textbf{$\Delta$}
& \textbf{Clean $\rightarrow$ Pert.} & \textbf{$\Delta$} \\
\midrule

GUI-Owl-1.5-2B
& $60.00 \rightarrow 20.00$ & \dec{-40.00}
& $25.00 \rightarrow 12.50$ & \dec{-12.50}
& $66.67 \rightarrow 25.00$ & \dec{-41.67} \\

GUI-Owl-1.5-4B
& $40.00 \rightarrow 0.00$ & \dec{-40.00}
& $62.50 \rightarrow 12.50$ & \dec{-50.00}
& $50.00 \rightarrow 25.00$ & \dec{-25.00} \\

GUI-Owl-7B
& $60.00 \rightarrow 20.00$ & \dec{-40.00}
& $75.00 \rightarrow 12.50$ & \dec{-62.50}
& $83.33 \rightarrow 58.33$ & \dec{-25.00} \\

GUI-Owl-1.5-8B
& $100.00 \rightarrow 20.00$ & \dec{-80.00}
& $62.50 \rightarrow 12.50$ & \dec{-50.00}
& $50.00 \rightarrow 41.67$ & \dec{-8.33} \\

GUI-Owl-32B
& $80.00 \rightarrow 20.00$ & \dec{-60.00}
& $75.00 \rightarrow 25.00$ & \dec{-50.00}
& $83.33 \rightarrow 66.67$ & \dec{-16.67} \\

GUI-Owl-1.5-32B
& $100.00 \rightarrow 0.00$ & \dec{-100.00}
& $56.25 \rightarrow 0.00$ & \dec{-56.25}
& $83.33 \rightarrow 66.67$ & \dec{-16.67} \\

\midrule
\textbf{Average}
& $73.33 \rightarrow 13.33$ & \dec{-60.00}
& $59.38 \rightarrow 12.50$ & \dec{-46.88}
& $69.44 \rightarrow 47.22$ & \dec{-22.22} \\

\bottomrule
\end{tabular}
}
\label{tab:most_influential_perturbation_types}
\end{table}

\begin{figure}
    \centering
    \includegraphics[width=\linewidth]{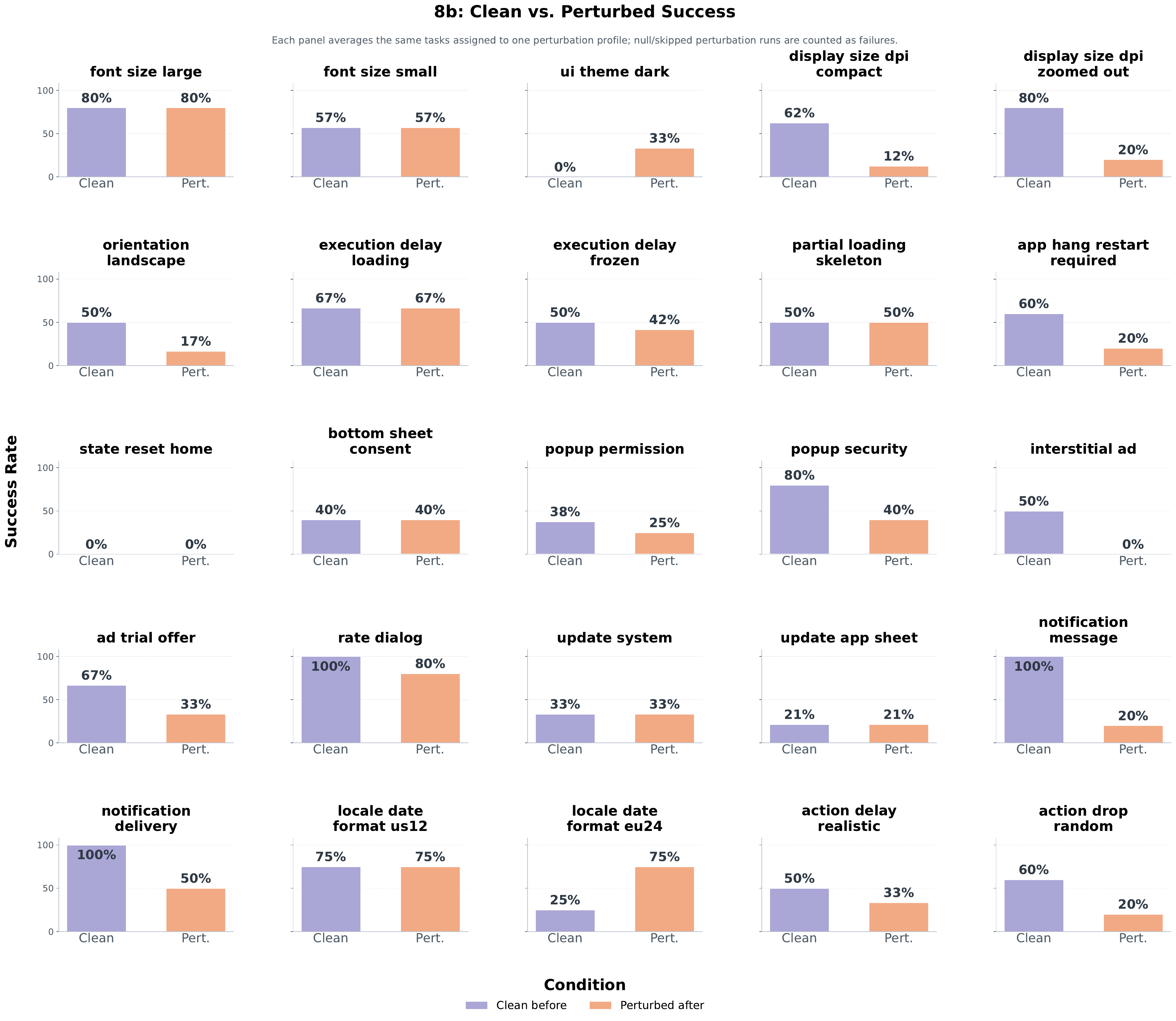}
    \caption{Clean-before vs.\ Perturbed-after success rates of GUI-Owl-1.5-8B across individual AndroidReality perturbation profiles. Following the GUI-Owl-1.5-32B analysis, this figure shows that the same perturbation-induced robustness gap also appears in the smaller 8B model, but with a different sensitivity profile. Several state-level perturbations continue to cause substantial degradation, especially display-density changes and landscape orientation, suggesting that the model remains highly dependent on canonical visual layouts and screen geometry. Interestingly, \texttt{ui\_theme\_dark} improves from the clean subset to the perturbed subset in this split, indicating that some perturbation-specific results can be affected by the underlying task subset and the model's baseline capability on those tasks, rather than reflecting uniformly harmful perturbations. Action-level perturbations show mixed behavior: execution loading, frozen execution, and partial loading are relatively stable, while app hangs and random action drops produce larger failures, suggesting that the model can tolerate some transient execution noise but struggles when the action effect is lost or the task context must be recovered. Transition-level perturbations again expose large variation: security pop-ups, interstitial ads, ad-trial offers, notification messages, and notification delivery lead to clear drops, while bottom-sheet consent, update-system, and update-app-sheet perturbations can be less disruptive.}
    \label{fig:GUIOwl8_result}
\end{figure}

\section{Test-Time Introspective Recovery}

\subsection{Detailed TTIR Strategies}
\label{app:ttir_routines}

This section provides the complete diagnostic signals and recovery prompts for each of the four TTIR strategies summarized in Section~\ref{sec:ttir_design}.

\paragraph{\ding{172} Stuck-loop detection $\Rightarrow$ alternative-action recovery.}
\label{app:ttir_cat1}

\textbf{Diagnosis.} Stuck-loop detection combines two complementary signals. 
First, the diagnoser inspects the recent action history and is asked to judge whether the agent is caught in a repeated unresolved attempt, returning \texttt{repeat\_failure\_detected} as a boolean. 
Second, as a deterministic safety net independent of the diagnoser's output, we additionally trigger this signal whenever the last $N$ steps (default $N{=}5$) emit identical action, summary, and thought. 
Either signal alone is sufficient to trigger recovery, which provides robustness against cases where the diagnoser fails to flag a clear loop.

\textbf{Recovery.} TTIR triggers an \emph{alternative-action} recovery: the recovery agent is explicitly informed that the previous coordinate has been repeatedly attempted without producing the intended effect, and is instructed to select a different visible target or take a different route to the same sub-goal, rather than retrying the failed coordinate. 
The recovery prompt carries the previously failed action together with the diagnosed reason, anchoring the recovery agent on the specific stuck pattern rather than re-planning the full task. 
This breaks the stale-coordinate loop by redirecting the agent toward an alternative grounding for the same intent.

\paragraph{\ding{173} Action-effect verification $\Rightarrow$ alternative-action recovery.}
\label{app:ttir_cat2}

\textbf{Diagnosis.} The diagnoser first infers the agent's previous \emph{intended effect} from the last action, its summary, and its thought. 
It then checks whether this intended effect is reflected in the current screenshot and assigns \texttt{previous\_intended\_effect\_satisfied} one of three values: \texttt{yes}, \texttt{no}, or \texttt{uncertain}. 
A value of \texttt{no} means that the action was issued, but the expected outcome is not visible on the screen. 
This indicates a likely silent execution failure, such as a dropped, delayed, or absorbed action.

\textbf{Recovery.} TTIR triggers an \emph{alternative-action} recovery: the recovery agent is prompted to attempt a different way to achieve the same intended effect, for example, instead of blindly reissuing the same action, it may select a different visible target, scroll to reveal an occluded element, or substitute an equivalent action. 
The prompt explicitly carries the previous intended effect as the recovery target, so that the recovery agent stays focused on the original sub-goal rather than re-planning the full task.

\paragraph{\ding{174} Interference Detection $\Rightarrow$ Dismiss-Interference Recovery.}
\label{app:ttir_cat3}

\textbf{Diagnosis.} The diagnoser inspects the screenshot for off-task overlays such as pop-ups, consent sheets, advertisements, system updates, and notification banners, and sets \texttt{interference\_present} accordingly. 
A value of \texttt{true} indicates that the agent should resolve the off-task element before continuing with the main task.

\textbf{Recovery.} TTIR triggers a \emph{dismiss-interference} recovery, with the explicit instruction to close, decline, or navigate past the interfering element. 
The recovery scope is restricted to interference resolution: the recovery agent is told that its only goal in this session is to make the original task UI visible again, and is forbidden from engaging with the content of the interruption (e.g., it does not click ``Sign up'' inside an ad-trial offer or proceed through a fake security wizard). 
Once the diagnoser confirms the screen is no longer blocked by the interference, the session terminates and the main agent resumes.

\paragraph{\ding{175} Task-Alignment Check $\Rightarrow$ Re-Orientation Recovery.}
\label{app:ttir_cat4}

\textbf{Diagnosis.} The diagnoser determines whether the current screen still belongs to the original task path, assigning \texttt{task\_alignment} one of \texttt{on\_track}, \texttt{off\_track}, or \texttt{uncertain}. 
This signal is particularly relevant after a successful interference dismissal, where the agent may have lost track of the original task and landed on a home screen, an app drawer, or an unrelated screen, indicating goal memory drift.

\textbf{Recovery.} TTIR triggers a \emph{re-orientation} recovery: the recovery agent is prompted with the original task goal and the most recent on-task subgoal, and is asked to issue the shortest action that restores the correct task context, for example, navigating back, re-opening the relevant app, or returning to the previously active screen. 
The recovery is again scoped: the agent does not attempt to complete the task, only to restore an on-task state from which the main agent can resume.

\subsection{Recovery Case Studies}
\label{app:recovery_cases}

In this section, we present qualitative examples in which each of the four recovery strategies discussed in Section~\ref{sec:ttir_design} successfully repairs a failed trajectory. As shown in Table~\ref{tab:mapping}, we summarize the mapping between the four recurring failure categories and the corresponding TTIR diagnosis and recovery strategies.

\begin{table}[t]
\centering
\small
\setlength{\tabcolsep}{4pt}
\renewcommand{\arraystretch}{1.15}
\caption{Mapping between the four recurring failure categories and the corresponding TTIR diagnosis and recovery strategies.}
\label{tab:ttir_mapping}
\resizebox{\linewidth}{!}{%
\begin{tabular}{p{0.51\linewidth} p{0.30\linewidth} p{0.30\linewidth}}
\toprule
\textbf{Failure Category} & \textbf{TTIR Diagnosis} & \textbf{Recovery Strategy} \\
\midrule

\textbf{\ding{172} Figure~\ref{fig:recovery_1}: Stale-Coordinate Stuck}
& \ding{172}  Stuck-loop detection
& Alternative-action recovery \\

\textbf{\ding{173}  Figure~\ref{fig:recovery_2}: Silent Action Failures}
& \ding{173}  Action-effect verification
& Alternative-action recovery \\

\textbf{\ding{174} Figure~\ref{fig:recovery_3}: Off-Task Engagement with Distractions}
& \ding{174} Interference detection
& Dismiss-interference recovery \\

\textbf{\ding{175}  Figure~\ref{fig:recovery_4}: Goal Memory Drift After Distraction}
& \ding{175}  Task-alignment check
& Re-orientation recovery \\

\bottomrule
\end{tabular}%
}
\label{tab:mapping}
\end{table}

\begin{figure}[h!]
    \centering
    \includegraphics[width=\linewidth]{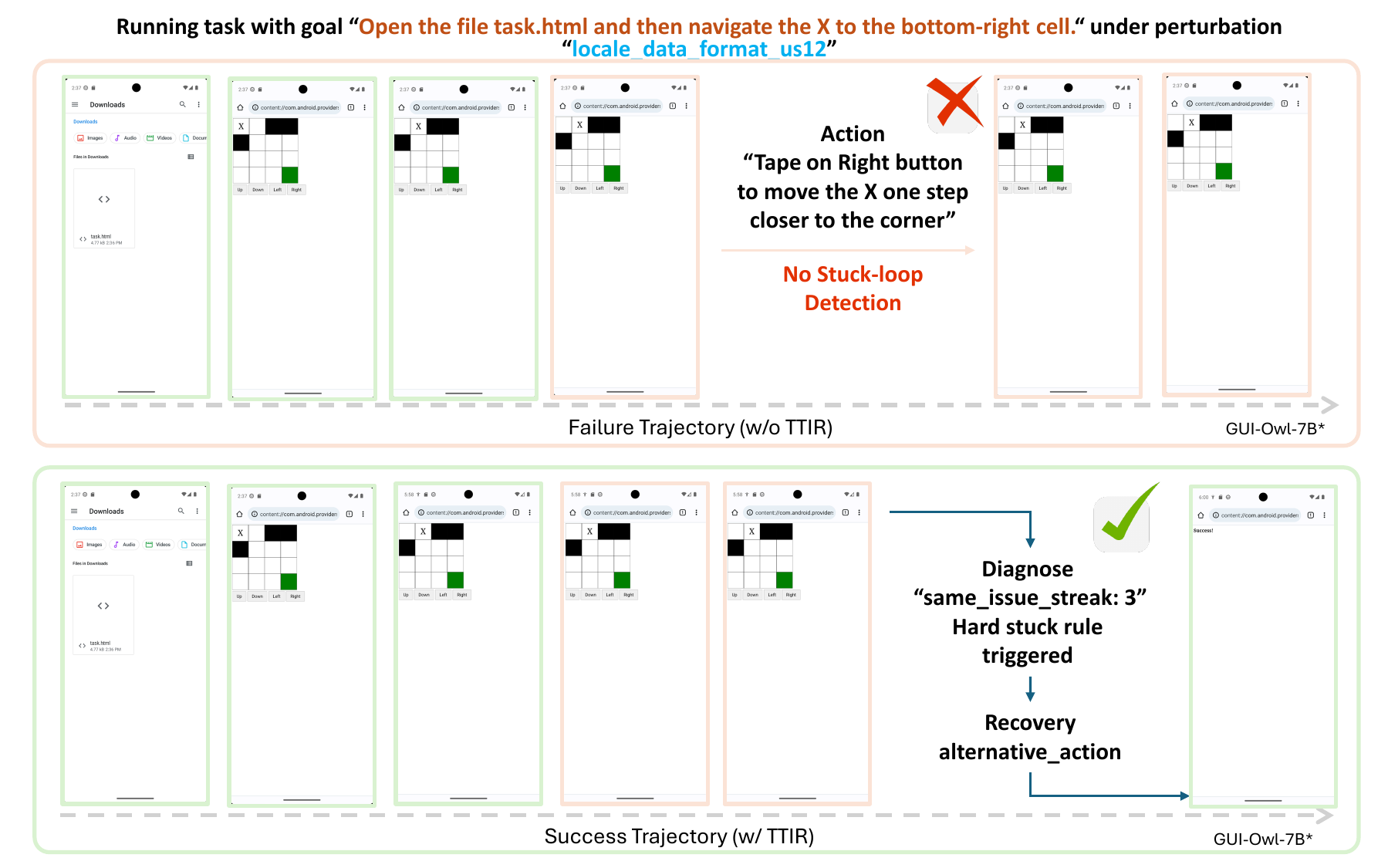}
    \caption{\textbf{\ding{172}Case study: stuck-loop recovery via TTIR.} 
    The task asks the agent to open \texttt{task.html} and navigate an ``X'' marker to the bottom-right cell of a grid using the on-screen \emph{Up}, \emph{Down}, \emph{Left}, and \emph{Right} buttons.
    \textbf{(Top, Failure trajectory without TTIR.)} 
    GUI-Owl-7B opens \texttt{task.html} correctly and identifies the \emph{Right} button as the appropriate move. 
    However, its predicted tap coordinate does not register on the actual button under the perturbed layout, and the X remains in place. 
    Without an explicit stuck-loop check, the agent continues to issue the same tap on the \emph{Right} button across consecutive steps, never re-grounding to a different visible target. 
    The trajectory exhausts the step budget and the X never reaches the bottom-right cell.
    \textbf{(Bottom, Success trajectory with TTIR.)} 
    With TTIR enabled, the same agent enters the same loop, but after three identical taps the diagnoser's hard-stuck rule fires (\texttt{same\_issue\_streak = 3}). 
    This triggers an \texttt{alternative\_action} recovery: the recovery agent is instructed that the previous coordinate has been repeatedly attempted without effect, and is asked to choose a different visible target or a different route to the same sub-goal. 
    Following the recovery action, the agent successfully advances the X and ultimately reaches the bottom-right cell, completing the task.}
    \label{fig:recovery_1}
\end{figure}

\begin{figure}[h!]
    \centering
    \includegraphics[width=0.99\linewidth]{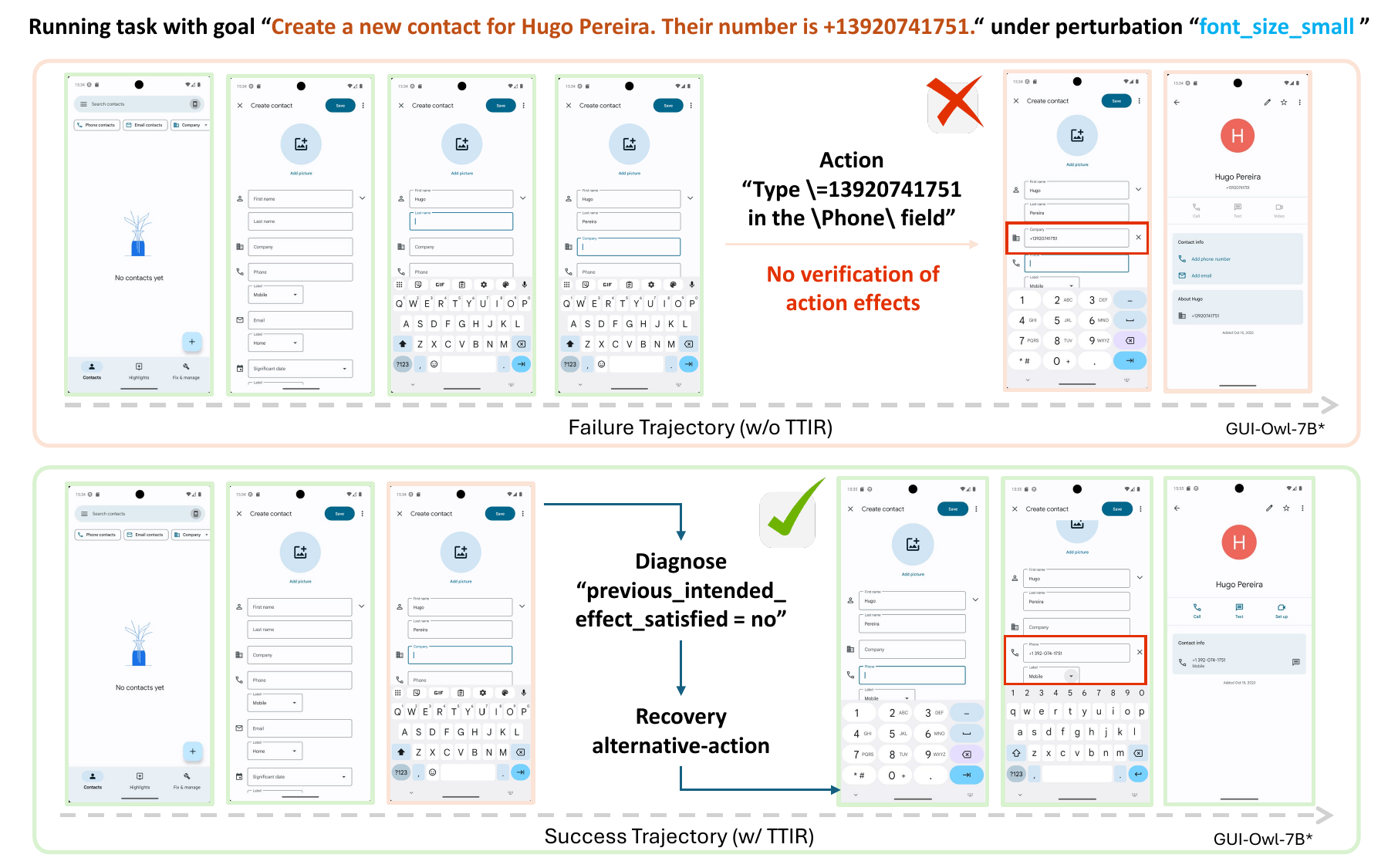}
    \caption{\textbf{\ding{173}Case study: silent execution failure recovered by TTIR.} 
    The task asks the agent to create a new contact for ``Hugo Pereira'' with phone number \texttt{+13920741751}, under the \texttt{font\_size\_small} state-level perturbation that compresses the form layout.
    \textbf{(Top, Failure trajectory without TTIR.)} 
    GUI-Owl-7B successfully fills in the \emph{First name} and \emph{Last name} fields, then attempts to type the phone number into the \emph{Phone} field. 
    However, because the perturbed font size shifts the relative positions of the form fields, the typed digits are silently entered into the \emph{Company} field instead, as highlighted by the red box. 
    Without an explicit verification step between issuing the action and continuing, the agent does not notice the misplacement and proceeds to save the contact, producing a final record where the phone number is stored under \emph{Company} rather than \emph{Phone}. 
    The task is judged as failed.
    \textbf{(Bottom, Success trajectory with TTIR.)} 
    With TTIR enabled, the same misplacement occurs initially, but before the agent's next action, the diagnoser compares the intended effect (``phone number entered in the Phone field'') against the current screenshot and sets \texttt{previous\_intended\_effect\_satisfied = no}. 
    This triggers an \texttt{alternative\_action} recovery: the recovery agent clears the misplaced text and re-enters the phone number into the correct \emph{Phone} field. 
    After the recovery completes, the main agent saves the contact with all fields correctly filled, and the task succeeds.}
    \label{fig:recovery_2}
\end{figure}

\begin{figure}[h!]
    \centering
    \includegraphics[width=0.99\linewidth]{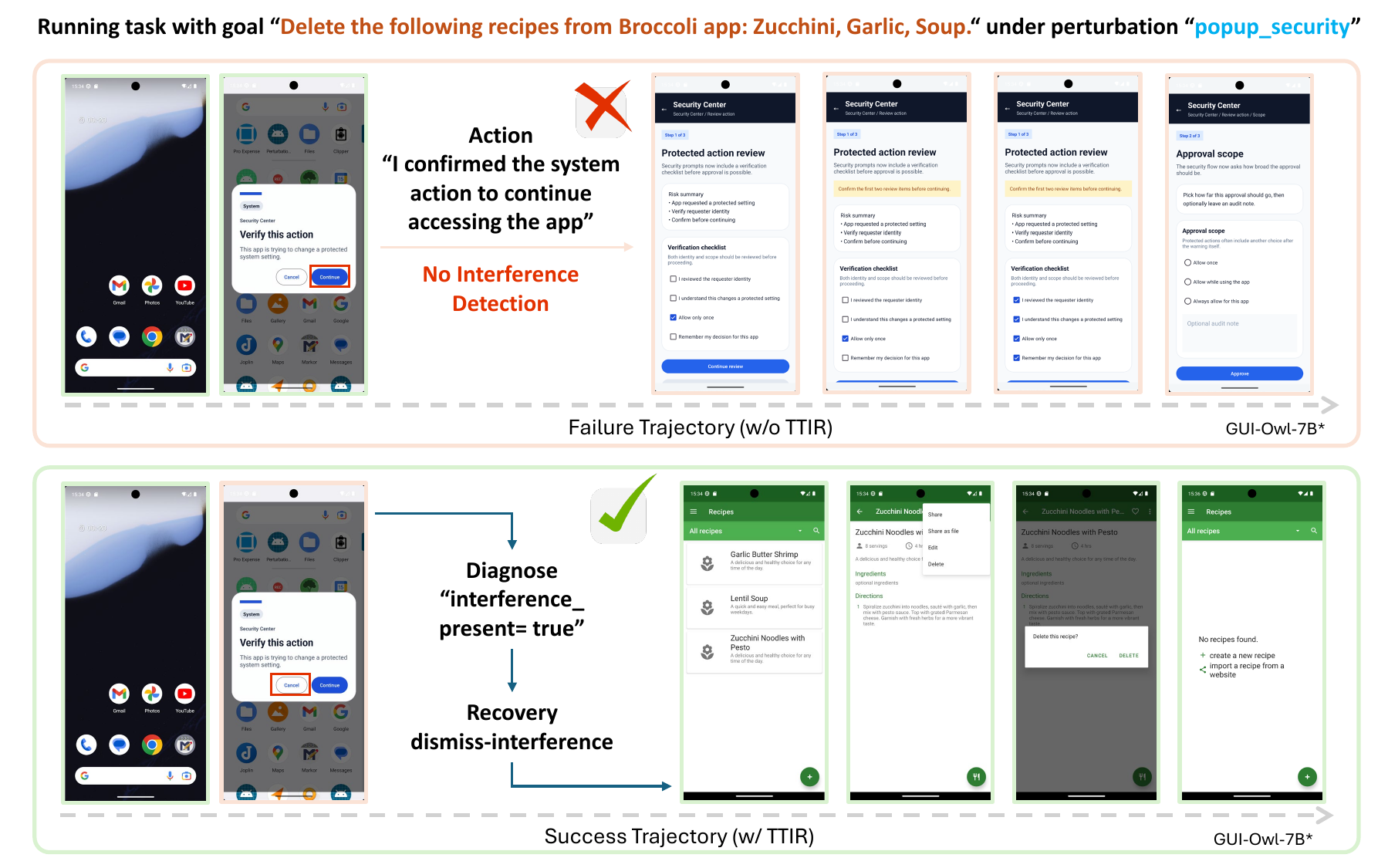}
    \caption{\textbf{\ding{174}Case study: off-task engagement with distractions via TTIR.} 
    The agent is asked to delete three recipes (Zucchini, Garlic, Soup) from the Broccoli recipe app, under the \texttt{popup\_security} transition-level perturbation that injects a fake ``Verify this action'' security pop-up upon launching the app.
    \textbf{(Top, Failure trajectory without TTIR.)} 
    GUI-Owl-7B encounters the security pop-up but fails to recognize it as off-task. 
    Instead of dismissing it, the agent clicks the \emph{Continue} button, treating the pop-up as part of the intended workflow. 
    This leads it into a multi-step ``Security Center / Protected action review'' wizard injected by the perturbation, where the agent dutifully reviews the risk summary, checks all verification checkboxes across three pages, and proceeds to the approval-scope screen. 
    By the time the wizard ends, the step budget is exhausted; the agent never reaches the Broccoli app, and not a single recipe is deleted.
    \textbf{(Bottom, Success trajectory with TTIR.)} 
    With TTIR enabled, the diagnoser inspects the same screen and identifies the security pop-up as off-task, setting \texttt{interference\_present = true}. 
    This triggers a \texttt{dismiss-interference} recovery: the recovery agent clicks the \emph{Cancel} button (red box) and the pop-up disappears, restoring the original task UI. 
    Control then returns to the main agent, which proceeds normally to open the Broccoli app, locate each target recipe, and delete them through the recipe-detail menu, completing the task successfully.
}
    \label{fig:recovery_3}
\end{figure}

\begin{figure}[h!]
    \centering
    \includegraphics[width=0.99\linewidth]{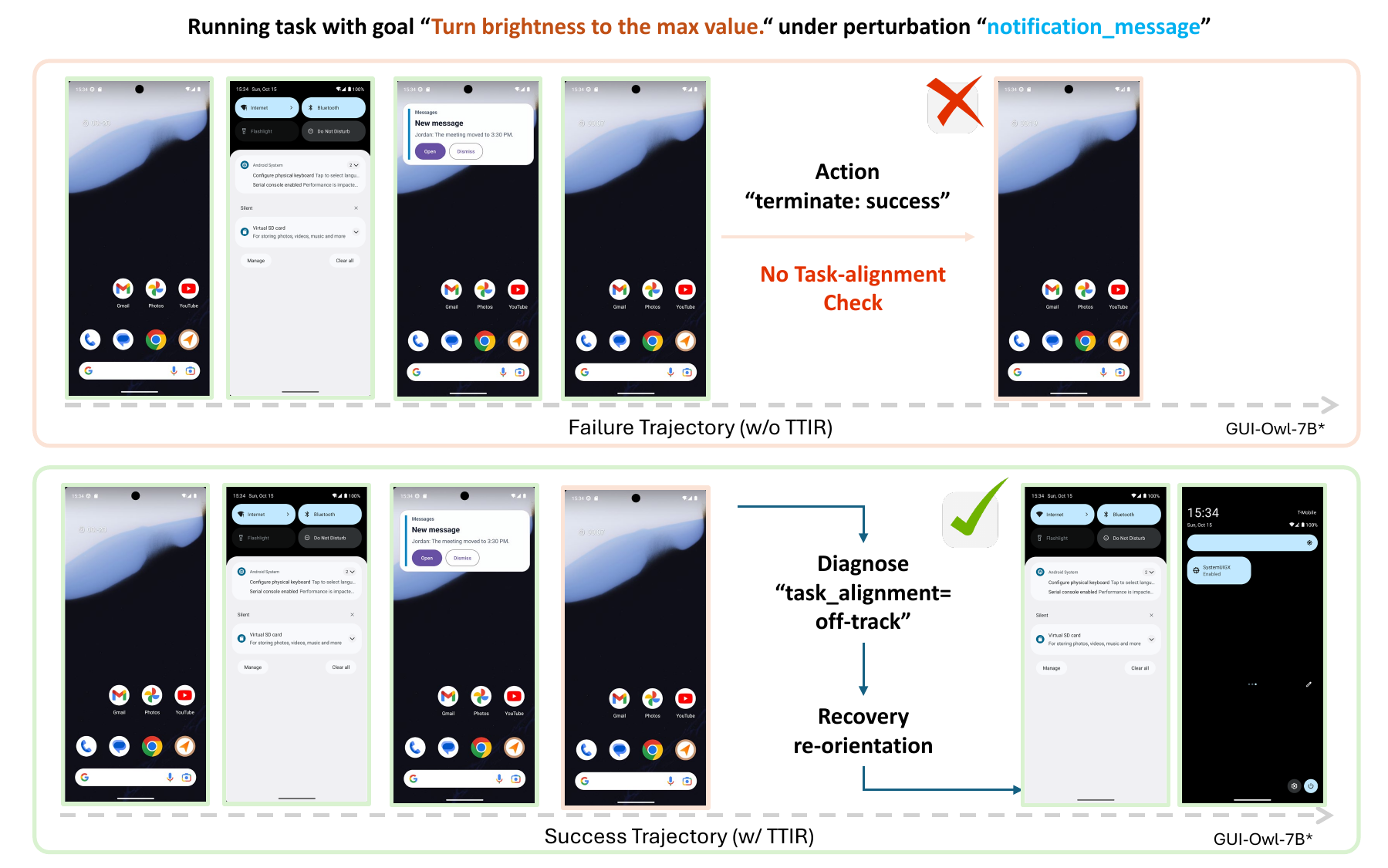}
    \caption{\textbf{\ding{175}Case study: goal drift detection via TTIR.} 
    The agent is asked to set the screen brightness to its maximum value, under the \texttt{notification\_message} transition-level perturbation that injects a ``New message'' notification banner partway through the trajectory.
    \textbf{(Top, Failure trajectory without TTIR.)} 
    GUI-Owl-7B begins correctly by pulling down the Quick Settings panel to access the brightness control. 
    A new message notification then appears and is dismissed by the agent. 
    However, after the dismissal, the agent finds itself back on the home screen and, lacking any check that this state is still consistent with the original task, concludes that the task is complete and emits \texttt{terminate(success)} without ever adjusting the brightness slider. 
    Although the agent successfully handled the distraction itself, the brief interruption was sufficient to disrupt its working memory of the original task plan.
    \textbf{(Bottom, Success trajectory with TTIR.)} 
    The same dismissal occurs, but TTIR's diagnoser then compares the current screen against the task goal and sets \texttt{task\_alignment = off-track}, recognizing that returning to the home screen does not satisfy a brightness-adjustment task. 
    This triggers a \texttt{re-orientation} recovery: the recovery agent re-opens the Quick Settings panel, locates the brightness slider, and drags it to the maximum value, restoring the trajectory to the original task path. 
    Control then returns to the main agent, which verifies the change and the task succeeds.}
    \label{fig:recovery_4}
\end{figure}

\end{document}